\documentclass[11pt]{article}
\usepackage[preprint]{acl}

\usepackage[T1]{fontenc}
\usepackage[utf8]{inputenc}
\usepackage{times}
\usepackage[expansion=false]{microtype}

\usepackage{amsmath, amssymb}
\usepackage{xcolor}
\usepackage{booktabs}
\usepackage{multirow}
\usepackage{tabularx}
\usepackage{enumitem}
\usepackage{float}
\usepackage{placeins}
\usepackage{url}
\usepackage{listings}
\usepackage{graphicx}
\usepackage{tikz}
\usetikzlibrary{positioning,arrows.meta,calc,shapes.geometric,fit,backgrounds}

\definecolor{codebg}{RGB}{245,245,245}
\newcommand{\systemname}{\textsc{MASDR-RAG}}
\newcommand{\hybridname}{\textsc{Hybrid-Routed}}
\newcommand{\numDocs}{1{,}128}
\newcommand{\numChunks}{88{,}907}
\newcommand{\numQueries}{200}

\title{When More Documents Hurt RAG: Measuring and Reducing Vector Search Dilution with Domain-Scoped Retrieval}

\author{Nabaraj Subedi$^*$, Ahmed Abdelaty, and Shivanand Venkanna Sheshappanavar\\
University of Wyoming, Laramie, WY 82071, USA\\
\{nsubedi1, aahmed3, ssheshap\}@uwyo.edu\\
$^*$Correspondence author}

\begin{document}
\maketitle

\begin{abstract}
Retrieval-augmented generation (RAG) degrades when applied to large, heterogeneous document collections.  In RAG systems, dense similarity becomes less discriminative, causing top-$k$ retrieval to return semantically similar but contextually incorrect chunks. We refer to this failure mode as \emph{vector search dilution}. Even under hybrid dense+sparse retrieval, we first observed it in a deployed Wyoming Department of Transportation (WYDOT) system; in a controlled experiment, growing the corpus from $54$ to \numDocs{} documents cut precision@10 by $63\%$ and recall of the answer-bearing document by $23\%$. Similar patterns were observed in six other DOTs. To diagnose this issue, we introduce MASDR-RAG (Metadata-Scoped Domain Retrieval for RAG) as a controlled testbed over \numQueries{} human-validated queries, five LLM backbones, six public benchmark corpora, and two index stacks. In MASDR-RAG, domain scoping is the load-bearing component. We found hard partitioning to be unsafe, whereas training-free \emph{soft} scoping is safe when the scope structure is real. Furthermore, we can discover these scopes without hand-curated metadata. However, multi-agent orchestration triggers a \emph{precision–faithfulness paradox} that survives format controls, prompt controls, and blind human evaluation. Based on these varied outcomes, our practical recommendation is simple: \emph{soft-scope first, then perform a single synthesis call}. 
\end{abstract}

\section{Introduction}
\label{sec:introduction}

Retrieval-augmented generation (RAG) has become the dominant pattern for grounding LLM outputs in external knowledge~\citep{lewis2020rag,guu2020realm,gao2024ragsurvey}. However, the standard embed–index–retrieve–generate pipeline scales poorly on regulated enterprise corpora that span thousands of heterogeneous documents~\citep{barnett2024seven,wu2025rag4cm}. As the corpus grows across categories, dense retrieval loses discriminative power. This happens even when the approximate nearest neighbor (ANN) index returns the true nearest neighbors~\citep{malkov2020hnsw, johnson2021faiss}: those neighbors are semantically related to the query but contextually irrelevant.

We call this failure mode \emph{vector search dilution}, and we define and measure it. We first saw it in the Wyoming Department of Transportation production chatbot. Holding $32$ human-labeled queries and their answer-bearing documents fixed, growing the index from $54$ to \numDocs{} documents cut precision@10 by $63\%$ ($.759{\to}.281$) and recall of the answer-bearing documents by $23\%$ ($.975{\to}.750$; \S\ref{sec:dilution}). The effect is not specific to WYDOT. A query-free proxy for the dilution factor $\delta$, computed from the corpus alone before any system is built, is larger for smaller categories across seven independently built state-DOT corpora ($\rho{=}{-}0.62$, $p{<}10^{-7}$), and the estimate remains stable as corpora are added.

We evaluate over \numQueries{}  human-validated queries, five LLM backbones besides the production Gemini model (Qwen2.5-7B, Llama-3-8B, Claude-Haiku-4.5, GPT-5-mini, DeepSeek-V3), two index stacks (FAISS and Neo4j HNSW), and six benchmark corpora built from public data (EnterpriseComposite-9, HotpotQA-distractor, MultiHop-RAG, NQ-Open, FinanceBench, MMLU-Pro). Across these evaluations, we find that scoping improves retrieval precision, but its effect on answers depends on how we apply it. \textbf{Hard} partitioning, which commits each query to a single scope, is unsafe: where the routing signal is weak, it lowers correctness by up to $21.8$ points. Training-free \textbf{soft} scoping, which fuses scoped and global retrieval, never significantly lowers correctness and helps where the scope structure is real. The scopes need not be curated by hand: clustering document embeddings recovers them well enough to route within $4$ points of the hand taxonomy.

Wrapping scoped retrieval in multi-agent orchestration has a less expected effect: retrieval precision rises, but answer faithfulness falls. This \emph{precision–faithfulness paradox} holds under format-isomorphism control and shared-prompt control, and in a blind human study, the orchestrated system is again rated less faithful ($\Delta{=}{+}0.17$, $p{=}0.005$). Neither answer format nor redundant evidence explains it; the evidence points to cross-source synthesis.

The prescription is as follows: \emph{soft-scope first, then synthesize once}, keep hard scoping for organizational corpora with a reliable routing signal, and use orchestration only for queries that need multi-step tools, such as version comparison. Throughout, \systemname{} is the instrument that isolates these effects, not the system we recommend. Our contributions are threefold:

\begin{enumerate}[leftmargin=*, itemsep=1pt, topsep=3pt]
    \item \textbf{Measuring dilution:} We define vector search dilution with a query-free proxy, show in a controlled experiment that precision falls as distractor documents are added, and replicate the size effect across seven independently built state-DOT corpora (\S\ref{sec:dilution}).

     \item \textbf{Safe and unsafe scoping:} We show that hard scoping is unsafe, introduce a training-free soft-scoping variant that never significantly lowers correctness, and recover scopes without curated metadata (\S\ref{sec:architecture}--\S\ref{sec:discovery}).

    \item \textbf{A human-validated paradox:} We isolate a precision–faithfulness paradox in multi-agent orchestration that holds under format, prompt, and blind human-evaluation controls (\S\ref{sec:paradox}).
\end{enumerate}

\section{Related Work}
\label{sec:related_work}

\paragraph{RAG and Dense Retrieval:}
RAG \citep{lewis2020rag} pairs a generator with a retriever and has since grown to include query transformation, re-ranking, and iterative retrieval \citep{gao2024ragsurvey}; \emph{agentic} variants \citep{singh2025agenticrag} let the model decide when to retrieve. Dense bi-encoders \citep{karpukhin2020dpr} and late-interaction models \citep{khattab2020colbert, santhanam2022colbertv2} have largely replaced sparse retrieval \citep{robertson2009bm25}, although hybrid schemes \citep{sawarkar2024blended} remain competitive on multi-domain corpora. Index scaling is usually treated as an algorithmic problem of approximate nearest-neighbor search \citep{malkov2020hnsw, johnson2021faiss}. Closer to our concern, prior work shows that dense retrieval loses discriminative power as the index grows \citep{reimers2021curse}, that irrelevant passages change what the generator produces \citep{cuconasu2024noise}, and that long contexts add noise \citep{jin2025longcontext}. Hubness explains why large categories crowd out small ones in nearest-neighbor lists \citep{radovanovic2010hubs}, and contextual chunk headers address this at indexing time \citep{anthropic2024contextual}. This line of work lacks a way to measure the problem from the corpus itself before building any system.

\paragraph{Query Routing and Multi-Agent Systems:}
\emph{Strategy routing} \citep{jeong2024adaptiverag, zhang2025ragrouter, guo2025routerag} picks a retrieval depth or a whole pipeline per query; it decides when and how hard to retrieve, but not where. \emph{Metadata filtering} \citep{poliakov2024multimetarag} masks candidates after retrieval while still searching the whole corpus. Neither uses the organizational scopes that already exist in the document graph to restrict the index at query time.

For orchestration, ReAct \citep{yao2023react} and LangChain \citep{chase2023langchain} provide general scaffolding for tool use. Genuinely multi-agent RAG assigns distinct roles that exchange messages: MA-RAG \citep{nguyen2025marag} chains task-specific agents, and SCOUT-RAG \citep{li2026scoutrag} runs cooperative domain-relevance and retrieval agents over graph domains. To our knowledge, no prior work examines how such coordination behaves once each tool returns cleaner, domain-scoped evidence. \systemname{} is deliberately simpler: a single reasoning agent with $K$ domain-scoped tools, so we can isolate the effect of orchestration.

\paragraph{Reranking, Iterative, and Graph RAG:} Two-stage pipelines rerank a bi-encoder top-$K$ with a cross-encoder \citep{nogueira2020monoT5}, and learned-sparse retrievers such as SPLADE \citep{formal2022splade,geng2024opensearchsparse} remain competitive. Iterative methods such as IRCoT \citep{trivedi2023interleaving}, Self-Ask \citep{press2023measuring}, and Self-RAG \citep{asai2024selfrag} share ReAct's multi-round loop, which becomes costly under open-source backbones. \citet{shi2023largelms} show that LLMs are distracted by irrelevant context, but they do not consider domain-precise context that is fragmented across sources. GraphRAG \citep{edge2024local} builds entity--relationship graphs from text and does not use the organizational metadata that enterprise document graphs already carry.

\paragraph{Evaluation:} RAGAS \citep{es2024ragas} provides standard retrieval and faithfulness metrics. Standard benchmarks such as Natural Questions \citep{kwiatkowski2019natural}, HotpotQA \citep{yang2018hotpotqa}, MultiHop-RAG \citep{tang2024multihoprag}, and long-context suites \citep{yen2025helmet} rely on homogeneous or synthetic corpora, so they lack the cross-domain dilution of a regulated enterprise collection. We still evaluate on them as boundary cases, and we assemble \textsc{EnterpriseComposite-9} to cover the multi-domain setting they miss.

\section{Vector Search Dilution}
\label{sec:dilution}

\subsection{System Context}

The corpus contains \numDocs{} documents spanning construction specifications, design manuals, materials testing procedures, crash reports, transportation improvement programs, and administrative reports, stored in Neo4j as \(\text{Document} \rightarrow \text{Section} \rightarrow \text{Chunk}\). The deployed system uses Gemini Embedding (3{,}072-d), an HNSW vector index, and a BM25 full-text index; we call this configuration the \emph{Gemini production stack}. Chunk counts vary widely across the nine categories (Table~\ref{tab:distribution}). Traffic \& Crash reports supply $34.8\%$ of all chunks from $1.9\%$ of the documents, and the mean number of chunks per document ranges by category from $26$ to $1{,}406$, a $54\times$ spread. This imbalance drives the dilution measured below.

\begin{table}[H]
\centering
\scriptsize
\setlength{\tabcolsep}{2.5pt}
\resizebox{\columnwidth}{!}{%
\begin{tabular}{l r r r c r r r}
\toprule
 & \multicolumn{3}{c}{\textbf{Category}} & & \multicolumn{3}{c}{\textbf{Agent scope}} \\
\cmidrule(lr){2-4}\cmidrule(lr){6-8}
\textbf{Category} & Docs & Chunks & \% & $\delta$ ($n$) & Docs & Chunks & Red.\% \\
\midrule
Standard Specs        & 2   & 2{,}519  & 2.8  & .43 (23) & 22      & 3{,}140  & 96.5 \\
Construction Manual   & 21  & 6{,}641  & 7.5  & .10 (21) & 21      & 6{,}641  & 92.5 \\
Materials Testing     & 6   & 2{,}180  & 2.5  & .22 (20) & 7       & 2{,}184  & 97.5 \\
Design Manual         & 23  & 1{,}405  & 1.6  & .53 (12) & 21      & 1{,}366  & 98.5 \\
Traffic \& Crashes    & 22  & 30{,}922 & 34.8 & .16 (7)  & 22      & 30{,}922 & 65.2 \\
STIP                  & 59  & 13{,}634 & 15.3 & .17 (9)  & 58      & 13{,}607 & 84.7 \\
Annual Reports        & 46  & 2{,}341  & 2.6  & .12 (4)  & 47      & 2{,}439  & 97.3 \\
Bridge Program        & 28  & 5{,}399  & 6.1  & .21 (12) & 45      & 8{,}076  & 90.9 \\
Other (general agent) & 921 & 23{,}866 & 26.8 & --       & 1{,}128 & 88{,}907 & 0.0 \\
\midrule
\textbf{Total} & \numDocs & \numChunks & 100 & & & & \textbf{90.4}$^{\dagger}$ \\
\bottomrule
\end{tabular}}
\caption{WYDOT categories: size, mean per-query dilution $\delta$ ($n$ queries), and the search space of each category's agent. Agent filters match related series (App.~\ref{app:filters}), so scope and category sizes can differ. ${}^{\dagger}$Weighted average reduction.}
\label{tab:distribution}
\end{table}

\subsection{Definition}

Let $\mathcal{C}{=}\{c_1,\dots,c_N\}$ be $N$ chunks partitioned into $K$ categories $\mathcal{C}_1,\dots,\mathcal{C}_K$, $e:\mathcal{C}\to\mathbb{R}^d$ an embedding, and $q$ a query targeting category $k^\star$. The top-$m$ retrieval set is $R_m(q) = \arg\max_{S\subseteq\mathcal{C},|S|=m}\sum_{c\in S}\mathrm{sim}(e(q),e(c))$. Dilution occurs when global precision is much lower than scoped precision:
\[
\delta(q,k^\star)
\;=\; 1 - \frac{P_{\text{global}}(q)}{P_{\text{scoped}}(q)},
\]
where $P_{\text{global}}(q)$ is the fraction of $R_m(q)$ that belongs to $k^\star$ when retrieval ranges over all of $\mathcal{C}$, and $P_{\text{scoped}}(q)$ is the same fraction when retrieval is restricted to $\mathcal{C}_{k^\star}$. Since $P_{\text{scoped}}{\approx}1$ by construction, $\delta \approx 1 - P_{\text{global}}$: the share of the global top-$m$ that comes from other categories. $\delta{=}0$ means no dilution, and $\delta\to 1$ means severe dilution.

\subsection{Empirical Measurements}

Small categories suffer the most (Design $\delta{=}0.53$; Specs $\delta{=}0.43$), while most large ones resist it (Construction Manual $\delta{=}0.10$; Table~\ref{tab:distribution}); Annual Reports is a small category with low dilution ($\delta{=}0.12$). On WYDOT alone, the Spearman correlation between $\log$ chunk count and mean $\delta$ over the eight scopable categories is $\rho{=}{-}0.60$ ($p{=}0.12$), which is underpowered at $n{=}8$. These per-query values need a labeled query set, so for the cross-corpus test, we use the query-free proxy $\hat\delta$ defined in \S\ref{sec:cross_dot}. Pooling every scopable category from the seven state-DOT corpora of \S\ref{sec:cross_dot} on the shared document\_series axis gives $\rho{=}{-}0.618$ ($p{=}4.1{\times}10^{-8}$, $n{=}65$). Pooling corpora of different sizes does not create the effect: correlating $\hat\delta$ with $\log$ share-of-corpus still gives $\rho{=}{-}0.510$ ($p{=}1.4{\times}10^{-5}$). The estimate is also stable as corpora are added ($-0.601$, $-0.590$, and $-0.618$ at three, six, and seven corpora).

 \paragraph{Adding documents while holding the queries fixed:} The correlations above compare categories of different sizes. To isolate corpus size, we take the $32$ WYDOT queries that carry labeled gold documents, keep their $16$ gold documents in the index, and add randomly drawn distractor documents up to the full corpus, with five draws per size; the queries, the embedder (BGE-M3), and $k{=}10$ stay fixed. A retrieved chunk counts as relevant only if it comes from the query's own gold documents. Precision falls at every step (Table~\ref{tab:dilution_curve}): from $54$ to \numDocs{} documents, it drops by $63\%$ relative ($.759{\to}.281$), and recall drops by $23\%$ ($.975{\to}.750$), even though every answer-bearing document stays in the index throughout. 
 

\begin{table}[t]
\centering
\small
\setlength{\tabcolsep}{6pt}
\begin{tabular}{rrcc}
\toprule
\textbf{Docs} & \textbf{Chunks} & \textbf{P@10}$^{\text{doc}}$ & \textbf{R@10}$^{\text{doc}}$ \\
\midrule
$16$      & $10{,}052$  & $.831$            & $1.000$ \\
$54$      & $\sim$$20$k & $.759_{\pm.049}$  & $.975_{\pm.026}$ \\
$100$     & $\sim$$44$k & $.672_{\pm.083}$  & $.938_{\pm.044}$ \\
$200$     & $\sim$$57$k & $.559_{\pm.036}$  & $.931_{\pm.041}$ \\
$400$     & $\sim$$86$k & $.472_{\pm.056}$  & $.906_{\pm.038}$ \\
$700$     & $\sim$$136$k& $.358_{\pm.027}$  & $.838_{\pm.041}$ \\
$1{,}128$ & $217{,}752$ & $.281$            & $.750$ \\
\bottomrule
\end{tabular}
\caption{Gold-document precision and recall at $k{=}10$ as distractors are added to the WYDOT index ($n{=}32$ queries; subscripts: s.d.\ over five draws).}
\label{tab:dilution_curve}
\end{table}

Geometrically, dilution is source confusion at retrieval time: the retriever returns neighbors from the wrong source, and $\delta$ measures how often. Scoping does not improve the embedder. It restricts the retrieval neighborhood to a source label that the document graph already holds, which helps when a query can be routed to the right label and hinders when it cannot. App.~\ref{app:tsne} shows this in a t-SNE projection, and App.~\ref{app:failure_case} walks through a WYDOT failure.

\subsection{Replication Across Seven State Agencies}
\label{sec:cross_dot}

A single organization could reflect WYDOT's own document practices. We therefore repeat the measurement on six more state DOTs: \textbf{California (Caltrans)}, \textbf{Colorado (CDOT)}, \textbf{Minnesota}, \textbf{Oregon}, \textbf{Texas}, and \textbf{Florida}. All were scraped from public portals and processed with one pipeline ($1000$-character chunks, BGE-M3; App.~\ref{app:cross_dot_appendix}).On each, we compute a query-free proxy for $\delta$: every chunk serves as a query, $P_{\text{global}}$ is the share of its $10$ nearest chunks from its own category, and $\hat\delta = 1 - P_{\text{global}}$ is averaged over the category. WYDOT is re-chunked for this comparison, so its values differ slightly from those above.

Small categories dilute more on the section axis in every corpus, and on the document\_series axis in all but Caltrans (Table~\ref{tab:dot_dilution}). Caltrans packages its Standard Specs as omnibus PDFs averaging $2{,}374$ chunks each (against $37$ for CDOT), each split into about $80$ topical sections; measured on sections, it reaches $\rho{=}{-}0.85$, in line with the others. Scope granularity must therefore match how the producer packages its documents. Dilution is a reproducible property of large, heterogeneous corpora, and $\hat\delta$ measures it before building any system.

\begin{table}[htbp]
\centering
\scriptsize
\renewcommand{\arraystretch}{0.99}
\begin{tabular}{l r r r r}
\toprule
Corpus & \multicolumn{2}{c}{document\_series} & \multicolumn{2}{c}{section} \\
\cmidrule(lr){2-3}\cmidrule(lr){4-5}
 & \#cat & $\rho$ & \#cat & $\rho$ \\
\midrule
CDOT     & $10$ & $-0.95$ & $208$ & $-0.89$ \\
WYDOT    & $10$ & $-0.69$ & $222$ & $-0.67$ \\
FDOT     &  $8$ & $-0.67$ & $393$ & $-0.61$ \\
ODOT     & $10$ & $-0.55$ & $479$ & $-0.82$ \\
TxDOT    &  $9$ & $-0.53$ & $286$ & $-0.85$ \\
MnDOT    &  $9$ & $-0.42$ &  $32$ & $-0.67$ \\
Caltrans &  $9$ & $-0.10^{\dagger}$ & $407$ & $-0.85$ \\
\bottomrule
\end{tabular}
\caption{Spearman $\rho$ between $\log$ category size and $\hat\delta$ for two scope axes (\#cat: number of categories); negative $\rho$ means smaller categories dilute more. ${}^{\dagger}$Caltrans use omnibus PDFs (only its \texttt{section} axis shows the effect).}
\label{tab:dot_dilution}
\end{table}

\section{Scoping Mechanisms}
\label{sec:architecture}

\begin{figure*}[t]
\centering
\scalebox{0.85}{%
\begin{tikzpicture}[
  font=\small,
  node distance=0.55cm and 1.25cm,
  box/.style   ={rectangle, draw, rounded corners, minimum height=0.75cm,
                 align=center, inner sep=4pt, font=\small},
  io/.style    ={box, fill=yellow!18, minimum width=2.05cm},
  router/.style={box, fill=orange!22, minimum width=2.55cm},
  orch/.style  ={box, fill=purple!12, minimum width=2.55cm},
  agent/.style ={rectangle, draw, rounded corners, fill=blue!9,
                 text width=2.55cm, align=center, font=\scriptsize,
                 minimum height=0.55cm, inner sep=2pt},
  agentEll/.style={font=\scriptsize\itshape, text=black!55},
  db/.style    ={cylinder, draw, fill=green!12,
                 shape border rotate=90, minimum height=1.05cm,
                 minimum width=1.7cm, aspect=0.32,
                 align=center, font=\scriptsize},
  synth/.style ={box, fill=red!8, minimum width=2.55cm},
  arrow/.style ={-{Latex[length=2mm]}, thick, draw=black!72},
  data/.style  ={font=\scriptsize\itshape, text=black!60, inner sep=1pt},
  group/.style ={draw=black!25, dashed, rounded corners,
                 inner sep=6pt, fill=blue!4},
]

\node[io]     (query)  {User\\Query};
\node[router] (router) [right=of query] {Hybrid Router\\(Regex $\to$ LLM)};
\node[orch]   (orch)   [right=of router] {Orchestrator\\(function-calling)};

\node[agent]  (a1) [right=1.4cm of orch, yshift= 3.2cm]
   {Specs Agent\\\textcolor{black!55}{$|\mathcal{C}|{\approx}3.1$k}};
\node[agent]  (a2) [below=0.18cm of a1]
   {Construction Agent\\\textcolor{black!55}{$|\mathcal{C}|{\approx}6.6$k}};
\node[agent]  (a3) [below=0.18cm of a2]
   {Materials Agent\\\textcolor{black!55}{$|\mathcal{C}|{\approx}2.2$k}};
\node[agent]  (a4) [below=0.18cm of a3]
   {Design Agent\\\textcolor{black!55}{$|\mathcal{C}|{\approx}1.4$k}};
\node[agent]  (a5) [below=0.18cm of a4]
   {Safety Agent\\\textcolor{black!55}{$|\mathcal{C}|{\approx}30.9$k}};
\node[agent]  (a6) [below=0.18cm of a5]
   {Bridge Agent\\\textcolor{black!55}{$|\mathcal{C}|{\approx}8.1$k}};
\node[agent]  (a7) [below=0.18cm of a6]
   {STIP Agent\\\textcolor{black!55}{$|\mathcal{C}|{\approx}13.6$k}};
\node[agent]  (a8) [below=0.18cm of a7]
   {Annual Reports Agent\\\textcolor{black!55}{$|\mathcal{C}|{\approx}2.4$k}};
\node[agent]  (a9) [below=0.18cm of a8]
   {General Agent\\\textcolor{black!55}{$|\mathcal{C}|{\approx}88.9$k}};

\node[db]     (kg) [right=2.4cm of a5.east] {Neo4j\\Knowledge\\Graph};

\node[synth]  (synth) [below=1.4cm of orch] {Synthesiser LLM\\\scriptsize(Qwen-7B / Llama-8B / Gemini)};
\node[io]     (ans)   [below=of synth] {Answer};

\draw[arrow] (query)  -- (router);
\draw[arrow] (router) -- node[data, above]{category} (orch);

\coordinate (bus) at ($(orch.east)+(0.55cm,0)$);
\draw[thick, draw=black!72] (orch.east) -- (bus);
\foreach \a in {a1, a2, a3, a4, a5, a6, a7, a8, a9} {
  \draw[arrow, rounded corners=2pt] (bus) |- (\a.west);
}

\coordinate (kgIn) at ($(kg.west)+(-0.45cm,0)$);
\foreach \a in {a1, a2, a3, a4, a5, a6, a7, a8, a9} {
  \draw[arrow, rounded corners=2pt] (\a.east) -| (kgIn) -- (kg.west);
}
\node[data] at ($(a9.south -| kgIn)+(0,-0.35cm)$) {scoped ANN search};

\coordinate (kgOut)   at ($(kg.north west)+(0.05cm,0)$);
\coordinate (corrAnchor) at ($(a1.north)+(0,0.45cm)$);
\coordinate (kgUp)    at (kgOut       |- corrAnchor);
\coordinate (orchUp)  at (orch.north  |- corrAnchor);
\draw[arrow, dashed, rounded corners=4pt]
  (kgOut) -- (kgUp) -- (orchUp) -- (orch.north);
\node[data, above=1pt] at ($(kgUp)!0.5!(orchUp)$) {top-$k$ chunks};

\draw[arrow] (orch.south) -- node[data, right]{chunks + query} (synth.north);
\draw[arrow] (synth.south) -- (ans.north);

\begin{pgfonlayer}{background}
  \node[group, fit=(a1)(a9)(kg),
        label={[font=\scriptsize\itshape, text=black!55, yshift=-2pt]below:scoped retrieval: $65$--$98\%$ search-space reduction}] {};
\end{pgfonlayer}

\end{tikzpicture}%
}
\caption{\systemname{} / \hybridname{} data flow. Regex-then-LLM router dispatches to one of eight domain agents or the general agent; each agent ANN-searches its document\_series scope in the Neo4j graph (sizes from Table~\ref{tab:distribution}), and a Qwen-7B / Llama-8B / Gemini synthesizer generates the answer.}
\label{fig:architecture}
\end{figure*}

We compare four mechanisms. (1) \textbf{Domain-scoped retrieval}: each agent restricts search to documents that match a Neo4j metadata filter, which shrinks its search space by $65$--$98\%$ (weighted average $90.4\%$; Table~\ref{tab:distribution}). (2) \textbf{Hybrid routing} (\hybridname{}): a fast regex matcher runs first and falls back to a zero-shot LLM classifier, so a query needs at most two LLM calls. (3) \textbf{Soft-scoping}, our recommended variant, is described below. (4) \textbf{Scope-routed orchestration} (\systemname{}): a single reasoning agent dispatches the query to $K$ domain-scoped tools through function calling, over up to five rounds; its per-domain ``agents'' are tool configurations, not separate agents. Figure~\ref{fig:architecture} shows the data flow, and Table~\ref{tab:methods} lists every system we compare.

\paragraph{Soft-scoping:} Hard routing commits each query to one scope, so a single mis-route loses the evidence. Soft-scoping avoids this by construction. A training-free \emph{centroid router} scores the query against each scope's mean chunk embedding, retrieves the top-$s$ scopes together with an unscoped global list, and merges the lists by reciprocal-rank fusion before one synthesis call. Because the global list always takes part, the worst case falls back to monolithic retrieval instead of the wrong partition.

\begin{table}[htbp]
\centering
\scriptsize
\renewcommand{\arraystretch}{0.92}
\resizebox{\columnwidth}{!}{%
\begin{tabular}{l p{5.4cm}}
\toprule
\textbf{System} & \textbf{Defining difference} \\
\midrule
Monolithic          & Global top-$k$ over the full index; one synthesis call \citep{lewis2020rag}. \\
Mono+RRF            & Monolithic with reciprocal-rank fusion \citep{cormack2009rrf} of dense $+$ sparse. \\
Regex-Scoped        & Keyword regex picks one scope; global fallback; one call. \\
LLM+Scoped          & Zero-shot LLM router picks one scope; one call. \\
\hybridname{}       & Regex fast-path $\to$ LLM router; one scoped call. \\
\textbf{Soft-Scoped} & \textbf{Centroid router; top-$s$ scope union $+$ global, RRF-fused; one call.} \\
\textsc{R2-Routed}  & Trained BGE-M3 \citep{bge_m3} linear-probe router; one scoped call. \\
\systemname{}       & Single reasoning agent, $K$ scope-bound tools, $\le$5 tool rounds. \\
\textsc{SingleCall} & \systemname{} retrieval, all evidence concatenated into one call. \\
ReAct               & Iterative reason–act loop over the same tools ($\le$6 rounds) \citep{yao2023react}. \\
MA-RAG              & Genuinely multi-agent: planner/extractor/QA chain \citep{nguyen2025marag} (port). \\
SCOUT-RAG           & Genuinely multi-agent: DRAA/PAGA/OASA/AQAA loop \citep{li2026scoutrag}. \\
\bottomrule
\end{tabular}}
\caption{Consolidated map of all systems compared, with each one's defining difference.}
\label{tab:methods}
\end{table}

\section{Evaluation on WYDOT: Retrieval Precision vs.\ Answer Quality}
\label{sec:evaluation}

We evaluate the WYDOT \numQueries{}-query suite, first on the Gemini production stack that motivated the study and then in an open-source replication. Significance uses paired permutation tests ($10{,}000$ permutations, $\alpha{=}0.05$).

\paragraph{Metrics:} We report P@10 and R@10 against each query's target scope, as well as binary correctness (Corr) and faithfulness (Faith) judged by Qwen-2.5-7B and averaged over queries. A Llama-3-8B re-judge ($792$ answers, $\kappa{=}0.59$) leaves the correctness rankings unchanged, and a human rater checks faithfulness (\S\ref{sec:paradox}).

\paragraph{Configurations:} Unless a caption says otherwise, results use Qwen-2.5-7B as synthesizer and judge, BGE-M3 embeddings, a FAISS store, and one shared, corpus-neutral answer prompt (\S\ref{sec:cross_domain}). Tables~\ref{tab:retrieval} and~\ref{tab:wydot_extbaselines} use the production store, and the appendix runs under the original prompt state; we never pool across configurations.

\begin{table}[htbp]
\centering
\scriptsize
\setlength{\tabcolsep}{3pt}\renewcommand{\arraystretch}{0.99}
\begin{tabular}{lcccc}
\toprule
\textbf{System} & \textbf{P@10} & \textbf{R@10} & \textbf{Corr\%} & \textbf{Faith} \\
\midrule
Monolithic    & $.77$ & $.93$ & $25.5$ & $.61$ \\
Mono+RRF      & $.75$ & $\mathbf{.96}$ & $27.0$ & $.58$ \\
LLM+Scoped    & $.85^*$ & $.86$ & $24.1$ & $.62$ \\
\systemname{} & $\mathbf{.86}^*$ & $.59^{**}$ & $\mathbf{33.5}$ & $.35^{**}$ \\
\hybridname{} & $.83$ & $.84$ & $24.5$ & $\mathbf{.62}$ \\
\bottomrule
\end{tabular}
\caption{WYDOT on the Gemini production stack (Gemini~2.5 Flash, Gemini embeddings, Neo4j HNSW $+$ BM25; $n{=}200$). Scoping raises P@10 while \systemname{} loses recall and faithfulness. ${}^*p{<}.05$, ${}^{**}p{<}.01$ vs.\ Monolithic.}
\label{tab:retrieval}
\end{table}

Scoping lifts P@10 from $.77$ to $.86$ on the production stack (Table~\ref{tab:retrieval}), but it does not improve correctness. No scoped system significantly beats Monolithic on WYDOT, either here or under the canonical configuration (Table~\ref{tab:cross_domain}). Scoping's demonstrated value on WYDOT is therefore retrieval precision, as correctness under strict binary judging depends as much on the synthesizer as on the judge.

\paragraph{Open-source replication:}
\label{sec:oss}
We re-ran the systems with Qwen2.5-7B-Instruct and Llama-3-8B-Instruct synthesizers on BGE-M3~\citep{bge_m3} embeddings. On Llama-3-8B, the single-call systems stay within $7$ correctness points of each other on every public corpus, while \systemname{}'s faithfulness collapses (App.~\ref{app:backbones_full}). Cost differs more than quality: ReAct averages $5.5$ LLM calls per WYDOT query on Llama-3-8B against $2.3$ on Qwen-7B (Figure~\ref{fig:efficiency}).

\section{Cross-Corpus Generalization: Hard vs.\ Soft Scoping}
\label{sec:cross_domain}

To test transfer beyond WYDOT, we replicate on public corpora. Every system uses one identical, corpus-neutral answer prompt, and each corpus has its own router configuration. Under unmatched prompts, Soft-Scoped trailed Monolithic on MultiHop by $9.4$ points despite $94.3\%$ chunk overlap (App.~\ref{app:prompt_control}).

\paragraph{Corpora:} \textbf{Composite-9} combines nine public sources that approximate enterprise documents ($17{,}994$ chunks; App.~\ref{app:composite}). \textbf{MultiHop-RAG}~\citep{tang2024multihoprag} and \textbf{MMLU-Pro} use their published splits, and Table~\ref{tab:cross_domain} repeats WYDOT under the same configuration. External baselines (ReAct, BM25, ColBERTv2, MA-RAG~\citep{nguyen2025marag}, SCOUT-RAG~\citep{li2026scoutrag}) and a HotpotQA replication leave the ranking unchanged (App.~\ref{app:external_baselines}).

\begin{table}[htbp]
\centering
\scriptsize
\setlength{\tabcolsep}{4pt}\renewcommand{\arraystretch}{0.99}
\begin{tabular}{l l c c}
\toprule
\textbf{Corpus} & \textbf{System} & \textbf{Corr.\%} & \textbf{Faith.} \\
\midrule
\multirow{6}{*}{WYDOT}
 & Monolithic        & 31.5 & 0.49 \\
 & Regex-Scoped      & 34.0 & 0.50 \\
 & \hybridname{} (hard) & 32.5 & 0.48 \\
 & Soft-Scoped       & 31.0 & 0.53 \\
 & \textsc{R2-Routed} & 28.5 & 0.50 \\
 & \systemname{}     & 30.5 & 0.33$^{\downarrow}$ \\
\midrule
\multirow{5}{*}{Composite-9}
 & Monolithic        & 78.2 & 0.77 \\
 & Regex-Scoped      & 77.3 & 0.77 \\
 & \hybridname{} (hard) & 56.4$^{\downarrow}$ & 0.56$^{\downarrow}$ \\
 & \textbf{Soft-Scoped} & \textbf{82.7}$^{\uparrow}$ & 0.69$^{\downarrow}$ \\
 & \systemname{}     & 62.2$^{\downarrow}$ & 0.59$^{\downarrow}$ \\
\midrule
\multirow{5}{*}{MultiHop-RAG}
 & Monolithic        & 69.0 & 0.51 \\
 & Regex-Scoped      & 68.4 & 0.48 \\
 & \hybridname{} (hard) & 66.4 & 0.43$^{\downarrow}$ \\
 & Soft-Scoped       & 68.8 & 0.48 \\
 & \systemname{}     & 52.6$^{\downarrow}$ & 0.24$^{\downarrow}$ \\
\midrule
\multirow{5}{*}{MMLU-Pro}
 & Monolithic        & 51.4 & 0.38 \\
 & Regex-Scoped      & 51.0 & 0.38 \\
 & \hybridname{} (hard) & 44.4$^{\downarrow}$ & 0.25$^{\downarrow}$ \\
 & Soft-Scoped       & 51.4 & 0.31$^{\downarrow}$ \\
 & \systemname{}     & 46.1$^{\downarrow}$ & 0.34 \\
\bottomrule
\end{tabular}
\caption{Correctness and faithfulness under the shared answer prompt (canonical configuration; $n{=}200$/$225$/$500$/$500$ for WYDOT/Composite-9/MultiHop/MMLU-Pro). ${}^{\uparrow}$/${}^{\downarrow}$: significantly above/below Monolithic ($p{<}0.05$).}
\label{tab:cross_domain}
\end{table}

\paragraph{Hard versus soft scoping:} Table~\ref{tab:cross_domain} separates the two mechanisms. \textbf{Hard} partitioning (\hybridname{}) is unsafe. It significantly lowers correctness on Composite-9 ($-21.8$ points, $p{<}0.001$) and MMLU-Pro ($-7.0$, $p{=}0.002$), and it never significantly helps. Both failures come from routing. Composite-9 queries never name their source, so a 7B router reaches only $33\%$ top-1 accuracy. MMLU-Pro's subject scopes offer no retrieval advantage. \textbf{Soft}-scoping never significantly lowers correctness. It significantly beats Monolithic on Composite-9 ($+4.5$, $p{=}0.041$), the corpus where hard scoping fails the most. On WYDOT, no scoped system differs significantly from Monolithic.

This safety holds for correctness, not faithfulness. Hard scoping significantly lowers faithfulness on all three public corpora. Soft-Scoped lowers it on Composite-9 ($0.69$ vs.\ $0.77$) and MMLU-Pro ($0.31$ vs.\ $0.38$). Orchestration (\systemname{}) trails Monolithic in correctness on all three public corpora. It also significantly lowers faithfulness on WYDOT, Composite-9, and MultiHop (\S\ref{sec:paradox}).

\paragraph{Multi-agent baselines on WYDOT:} Table~\ref{tab:wydot_extbaselines} adds the two published multi-agent systems, which ran on the BGE-M3 Neo4j store. On that store, MA-RAG reaches only $11.0\%$ correctness, compared to $15.1$--$30.3\%$ for the single-call systems (Table~\ref{tab:singlecall}). SCOUT-RAG reaches $24.1\%$, within that range. Both make far more LLM calls, about $22$ and $10$ times as many as Regex-Scoped. Judged with the same judge and chunk-text database, both are also far less faithful ($0.07$ and $0.10$, compared to $0.22$--$0.37$; App.~\ref{app:external_marag_scout}).

\begin{table}[t]
\centering
\footnotesize
\begin{tabular}{l r r r}
\toprule
System & Corr.\% & Faith.\ & calls \\
\midrule
Monolithic       & $33.5$ & $0.61$ & $1.0$ \\
Regex-Scoped     & $35.1$ & $0.61$ & $1.0$ \\
\hybridname{}    & $30.3$ & $0.43$ & $1.5$ \\
\systemname{}    & $28.7$ & $0.48$ & $2.1$ \\
ReAct            & $24.5$ & $0.59$ & $2.3$ \\
\midrule
MA-RAG           & $11.0$ & $0.07^{\ddagger}$ & $22.3$ \\
SCOUT-RAG        & $24.1$ & $0.10^{\ddagger}$ & $10.4$ \\
\bottomrule
\end{tabular}
\caption{WYDOT baselines (Qwen-2.5-7B synthesizer; upper block on the Gemini-embedding store, original prompt, $n{=}198$). ${}^{\ddagger}$BGE-M3 store; compare with Monolithic's $0.35$ there. Calls: mean LLM calls per query.}
\label{tab:wydot_extbaselines}
\end{table}

\section{Scope Discovery Without Curated Metadata}
\label{sec:discovery}

Everything so far assumes a producer-defined taxonomy in the document graph. This is the most common objection to scoping: if scopes must be curated by hand, only organizations that maintain one can use the method. Clustering document embeddings with $k$-means removes this requirement. On WYDOT, chunk purity is $0.78$, and the two categories that drive dilution, Traffic and STIP, are recovered at $89\%$ and $86\%$.

A router trained on the discovered scopes reaches $74.8\%$ accuracy under $5$-fold cross-validation on the $155$ labeled queries, compared to $78.7\%$ with the hand taxonomy. End-to-end, correctness does not differ significantly ($\Delta{=}0.03$, $p{=}0.30$; App.~\ref{app:discovery}). Domain scoping is therefore open to tenants without a curated taxonomy at a fixed granularity.

Choosing the \emph{granularity} automatically is harder. Across the seven corpora, the alignment between unsupervised clusters and a candidate axis does not predict which axis resolves dilution ($\rho{=}0.21$, $p{=}0.65$). Granularity therefore remains a corpus-level heuristic based on chunks per document (\S\ref{sec:cross_dot}), which is sufficient to configure each tenant in a multi-tenant deployment.

\section{The Precision--Faithfulness Paradox and Its Causes}
\label{sec:paradox}

On the Gemini production stack, \systemname{} raises P@10 from $0.77$ to $0.86$ but lowers faithfulness from $0.61$ to $0.35$ (Table~\ref{tab:retrieval}). The prompt confound does not explain this. Under the shared prompt, \systemname{} is the least faithful system on WYDOT ($0.33$ vs.\ $0.49$ for Monolithic; $p{=}0.0007$). It also performs significantly worse than Monolithic on Composite-9 and MultiHop (Table~\ref{tab:cross_domain}). The drop is borderline on MMLU-Pro ($p{=}0.050$) and absent on FinanceBench. It also depends on configuration: under the original prompt on the BGE-M3 store, \systemname{} was slightly more faithful than Monolithic (Table~\ref{tab:singlecall}). Table~\ref{tab:paradox_causes} pairs each candidate cause with the intervention that tests it (full ablations in App.~\ref{app:paradox_ablations}).

\begin{table*}[t]
\centering
\small
\setlength{\tabcolsep}{6pt}\renewcommand{\arraystretch}{1.2}
\begin{tabularx}{\textwidth}{@{}l >{\raggedright\arraybackslash}X >{\raggedright\arraybackslash}X l@{}}
\toprule
\textbf{Candidate cause} & \textbf{Intervention} & \textbf{Result} & \textbf{Verdict} \\
\midrule
Routing noise        & Trained R2 router, overall top-1 $.471{\to}.755$ & Faith $.369$ vs.\ $.347$ (Monolithic) & ruled out \\
Within-scope ranking & Cross-encoder rerank (App.~\ref{app:rerank}) & Faith moves both ways (Mono, Hybrid) & ruled out \\
Retriever family     & SPLADE for BGE-M3 (App.~\ref{app:splade}) & No systematic winner & ruled out \\
Index implementation & FAISS vs.\ Neo4j HNSW (App.~\ref{app:infra_parity}) & $|\Delta|{\le}.094$; no correctness flips & ruled out \\
\addlinespace[2pt]
Context fragmentation & All evidence in \emph{one} synthesis call (App.~\ref{app:singlecall_full}) & Faith $.591{\to}.600$ ($p{=}.91$) & falsified \\
Answer format         & Monolithic-isomorphic re-synthesis & Faith $.391{\to}.279$; falls further & falsified \\
Evidence redundancy   & Near-duplicate filter, cosine $>0.95$ & Faith $.279{\to}.254$; no recovery & falsified \\
\addlinespace[2pt]
Backbone class        & Matched prompts, Qwen-2.5-7B on WYDOT & Faith $.49{\to}.33$ ($p{=}.0007$) & not a boundary \\
\bottomrule
\end{tabularx}
\caption{Candidate causes of the precision--faithfulness paradox, each paired with the intervention that tests it. Grouped by verdict. Full ablations in App.~\ref{app:paradox_ablations}.}
\label{tab:paradox_causes}
\end{table*}

The four retrieval-side explanations are ruled out: routing, within-scope ranking, retriever family, and index implementation leave faithfulness essentially unchanged. The next three are falsified because removing each suspected cause leaves faithfulness flat or lower. Collapsing orchestration into one call leaves faithfulness unchanged ($p{=}0.91$) and raises correctness by $8.4$ points ($p{=}0.022$), so the multi-round schedule costs answer quality without improving grounding. The explanation most consistent with what remains is cross-source synthesis. Two threats remain: the effect could belong to our orchestrator or to the automatic judge.

\paragraph{Better routing does not remove the drop:} The production regex routes only $47.1\%$ of WYDOT queries correctly. A trained BGE-M3 probe (R2) raises this to $.755$, reaching $94.7\%$ on single-domain and $95.5\%$ on section-lookup queries; its remaining errors fall on cross-domain and version queries, whose gold route is global anyway (App.~\ref{app:router_variants}). Yet R2-Routed's faithfulness stays close to Monolithic ($.369$ vs.\ $.347$; Table~\ref{tab:singlecall}). A confidence-gated fallback to global retrieval only lowers correctness to the Monolithic level, so it bounds the worst case but does not help on average.

\paragraph{The effect is not specific to our orchestrator:} If \systemname{} were simply a poor orchestrator, better-designed systems should avoid the drop. Yet two independently designed systems, MA-RAG~\citep{nguyen2025marag} and SCOUT-RAG~\citep{li2026scoutrag}, do worse. On the BGE-M3 store, they fall below \systemname{} on correctness ($11.0\%$ and $24.1\%$ vs. $27.4\%$) and faithfulness ($0.07$ and $0.10$ vs. $.391$; Table~\ref{tab:singlecall}), while making $10\times$ and $5\times$ as many LLM calls.

\paragraph{A blind human rating agrees on the direction:} The last threat is that the LLM judge, not people, penalizes orchestrated answers. An annotator rated $78$ answers to $39$ paired WYDOT questions, seeing each with its retrieved evidence but not its system, on a three-level scale that gives partial credit. The annotator also rated \systemname{} less faithful than Monolithic ($0.41$ vs.\ $0.58$; $p{=}0.005$). The judge overstates the gap ($0.56$ vs.\ $0.17$), and per-answer agreement is weak ($r{=}0.13$). We therefore read faithfulness as an ordinal, within-judge signal: it shows which system is more faithful, not by how much.

\begin{figure*}[t]
\centering
\includegraphics[width=0.85\textwidth]{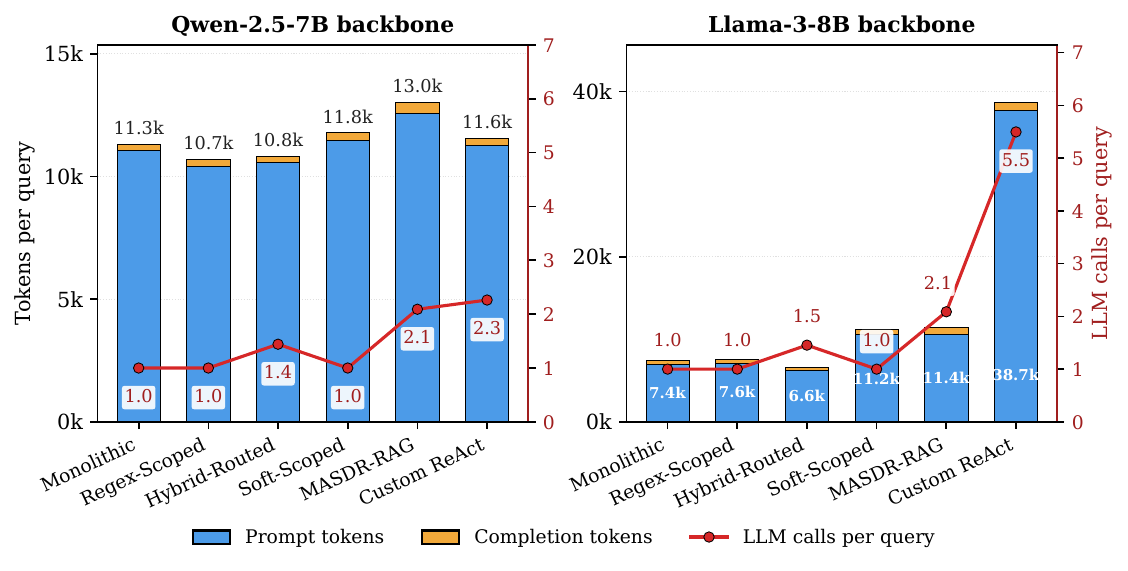}
\caption{Per-query tokens (stacked bars) and LLM calls (red line) on WYDOT ($n{\approx}200$). With Qwen-7B, all systems use $10.7$--$13.0$k tokens and at most $2.3$ calls; with Llama-8B, ReAct grows to $5.5$ calls and $38.7$k tokens, $5.9\times$ \hybridname{}'s budget.}
\label{fig:efficiency}
\end{figure*}

\section{Discussion: When to Scope}
\label{sec:discussion}

\paragraph{Cost:}
\label{sec:efficiency} Single-call scoping is also much cheaper than iterative agentic retrieval. On Llama-3-8B, ReAct reaches its $6$-iteration cap on half of WYDOT queries (mean $5.5$ calls against $1.5$ for \hybridname{}), using $5.9\times$ more tokens and $2.2\times$ the median latency (Figure~\ref{fig:efficiency}; App.~\ref{app:react}). With Qwen-7B's native function-calling template, the gap mostly closes, consistent with work on models trained for tool use~\citep{schick2023toolformer, parisi2022talm}.

\paragraph{When to scope:} Domain scoping pays only when three conditions hold. \textbf{(i) Meaningful organizational metadata}, curated or discovered (\S\ref{sec:discovery}), at a granularity that matches how documents are packaged (\S\ref{sec:cross_dot}). \textbf{(ii) Mostly single-scope queries}: when a query needs evidence from several scopes, partitioning removes the context it needs, as in MultiHop-RAG. \textbf{(iii) Reliable routing}: a router that cannot find the gold scope makes hard scoping harmful; on Composite-9, a 7B router reaches only $33\%$ top-1 accuracy, and hard scoping loses $21.8$ points.

The architectural choice follows. Where the routing signal is weak or unknown, \textbf{Soft-Scoped} is the safe default: it never significantly lowers correctness and wins where scope structure is real (Table~\ref{tab:cross_domain}). Where a reliable organizational signal exists, a trained router (\textsc{R2-Routed}) gives the highest WYDOT Recall@10 in the original-prompt BGE-M3 block ($.375$ against $.188$ for Monolithic; Table~\ref{tab:singlecall}) at half of \systemname{}'s LLM calls; although, under the shared prompt, it does not beat Monolithic on correctness ($28.5\%$ vs.\ $31.5\%$, not significant; Table~\ref{tab:cross_domain}). Where (i) or (ii) fails, stay monolithic. Orchestration is worth its cost only for queries that need multi-step tools, such as version comparison (App.~\ref{app:case_studies}). Scoping is the most consistent lever on retrieval precision, and the architecture built on top of it decides whether that gain reaches the answer.

\section{Conclusion}
We defined \emph{vector search dilution}, showed in a controlled experiment that retrieval precision falls as documents are added, and replicated the size effect across seven independently built state-DOT corpora ($\rho{=}{-}0.618$, $p{=}4.1{\times}10^{-8}$). Without metadata, clustering recovers the scopes. Under matched prompts, the mechanism of scoping decides the outcome: hard partitioning is unsafe, while soft partitioning never significantly lowers correctness and helps when the scope structure is real. Orchestration shows a precision–faithfulness paradox that holds under the shared-prompt and format controls, as well as in blind human evaluation. Where a corpus supports scoping, we recommend \emph{soft-scoping first and synthesizing once}.

\section*{Limitations}
\label{sec:limitations}

First, faithfulness rests on an LLM judge (Qwen-7B) that is also the synthesizer in most runs. A blind human study confirms the paradox's direction but shows that the judge overstates its size and uses a single annotator. Second, scoping's precision gain comes at the cost of recall, which a larger retrieval budget only partially recovers (App.~\ref{app:router_variants}). Third, our matched-prompt evidence covers only $7$--$8$B open-source models; commercial backbones ran at small sample sizes, and models of $70$B parameters or more are untested. Finally, we can discover scopes automatically, but we cannot yet choose their granularity without a heuristic.

\section*{Ethics Statement}
We use public government documents and public datasets. The system assists with document retrieval and does not generate policy recommendations. We used AI-assisted writing tools, including ChatGPT and Grammarly, to improve the manuscript's readability and grammatical clarity.

\section*{Reproducibility}
\label{sec:repro}
Code, the \numQueries{}-query WYDOT suite, all $6$ open-corpus ingest/eval scripts, SLURM submission scripts, and the judged per-query evaluation logs are released at \url{https://anonymous.4open.science/r/masdr-rag-E62C}; each table maps to one judged log file and one analysis script. \texttt{make reproduce} regenerates all tables CPU-only from the released logs, and the full GPU sweep re-runs on a single L40S. All models are public HuggingFace checkpoints; commercial backbones use OpenRouter. Full hyperparameters, hardware, and seeds in App.~\ref{app:reproducibility}.

\bibliography{custom}

\appendix
\renewcommand{\thesection}{A\arabic{section}}
\setcounter{section}{0}
\let\appdoldsection\section

\section{Data and Corpora}
\label{app:data}

This appendix describes the WYDOT query suite, the WYDOT agent filters, EnterpriseComposite-9, and the seven state-DOT corpora used in \S\ref{sec:dilution} and \S\ref{sec:cross_domain}.

\subsection{WYDOT Query Suite}
\label{app:query_validation}

The \numQueries{}-query suite was built in three passes. First, two authors drafted realistic operator questions for each of the nine WYDOT corpus sections, each with a reference answer and a gold document title. Next, a third author, not involved in drafting, removed ambiguous queries and queries without a clear answer in the corpus. This author also balanced the query types: single-domain ($113$), cross-domain ($31$), section-lookup ($22$), version-comparison ($27$), and ambiguous ($7$). Finally, we added queries that failed under the production system at the time (e.g.\ ``2020 construction manual'' against the larger 2021 corpus), so that systems cannot win on easy queries alone. Inter-author agreement on filter decisions was $89\%$ ($\kappa{=}0.74$, $40$-query sample). A $57$-query pilot gave the same system ordering as the full suite, with every core metric within $0.11$.

\subsection{WYDOT Agent Filters}
\label{app:filters}

Each agent's scope is defined by patterns on document\_series (Table~\ref{tab:agent_filters}), which run as Cypher \texttt{WHERE} clauses. A pattern can also match related series, so some scopes differ from their literal category in Table~\ref{tab:distribution}. For example, the Specs agent also matches Standard Plans. The router's categories are listed with the routing prompt (App.~\ref{app:routing_prompt}).

\begin{table}[!htbp]
\centering
\footnotesize
\resizebox{\columnwidth}{!}{%
\begin{tabular}{l l}
\toprule
\textbf{Agent} & \textbf{Filter (regex on document\_series)} \\
\midrule
specs        & \texttt{(?i).*standard.spec.*} \\
construction & \texttt{(?i).*construction.manual.*} \\
materials    & \texttt{(?i).*materials.*test.*} \\
design       & \texttt{(?i).*design.manual.*} \\
safety       & \texttt{(?i).*(crash|safety|fatal).*} \\
bridge       & \texttt{(?i).*bridge.*(prog|plan).*} \\
planning     & \texttt{(?i).*(stip|improv.prog).*} \\
admin        & \texttt{(?i).*(annual.rep|strateg).*} \\
general      & (no filter; full corpus) \\
\bottomrule
\end{tabular}}
\caption{Agent document\_series regex filters applied as Cypher \texttt{WHERE} clauses.}
\label{tab:agent_filters}
\end{table}

\subsection{EnterpriseComposite-9}
\label{app:composite}

EnterpriseComposite-9 combines nine public HuggingFace datasets, capped at $1{,}000$ documents per source (Table~\ref{tab:composite_sources}). Each chunk's source type serves as its routing label. GitHub and Jira both come from GitHub issue data, such as general issues and bug reports. Queries are generated with Qwen2.5-7B from the template below, and correctness is judged against the gold passage:
\begin{quote}\small
\texttt{Given the following \{source\_type\} passage, write a question whose answer is grounded in this passage and that a user of \{source\_type\} would plausibly ask.}
\end{quote}

\begin{table}[!htbp]
\centering
\footnotesize
\begin{tabular}{l l r}
\toprule
\textbf{Source} & \textbf{Dataset} & \textbf{Chunks} \\
\midrule
Confluence    & Wikipedia (en, sub.) & $2{,}354$ \\
Docs          & MS-MARCO passages    & $1{,}005$ \\
Gmail         & Enron email          & $1{,}847$ \\
Helpdesk      & HelpSteer            & $4{,}257$ \\
StackOverflow & Stack Overflow QA    & $1{,}849$ \\
Slack         & OpenAssistant chat   & $2{,}618$ \\
Github        & GitHub issues        & $1{,}519$ \\
Jira          & GitHub bug issues    & $1{,}508$ \\
Reports       & SEC filings (10-K)   & $1{,}037$ \\
\midrule
\textbf{Total} & & $\mathbf{17{,}994}$ \\
\bottomrule
\end{tabular}
\caption{Composite-9 composition (all nine source types ingested).}
\label{tab:composite_sources}
\end{table}

\subsection{Seven State-DOT Corpora}
\label{app:cross_dot_appendix}

\paragraph{Collection:} A single-threaded crawler fetched each DOT site to depth $4$, with at most $800$ HTML pages and $450$ PDFs per site. It paused $1$--$1.5$~s between requests, sent a descriptive user-agent with a contact email, and respected \texttt{robots.txt}. PDFs were deduplicated by content hash. Each received a provisional document\_series from URL and filename keywords, which a Qwen-2.5-7B classifier then refined; unclassified documents are tagged as General.

\paragraph{Chunking and embedding:} All corpora use $1000$-character chunks with $100$-character overlap and $L_2$-normalized BGE-M3 vectors. For WYDOT, we reconstructed each document from the production chunks and re-split it, increasing the chunk count from \numChunks{} to $217{,}752$.

\paragraph{Caltrans:} Caltrans Standard Specs consists of three yearly editions of one omnibus specification ($2023$--$2025$, about $4{,}600$--$4{,}900$ chunks each) and three short digests. Each edition is organized into about $80$ numbered sections. Within these $14{,}244$ chunks, smaller sections dilute more ($\rho{=}{-}0.79$): the largest, \textsc{section 96} ($1{,}202$ chunks), has $\delta{=}0.03$, while sections under $100$ chunks have $\delta{\ge}0.20$. The size effect thus holds even inside a single document, and sub-scoping by section shrinks the Specs search space to at most $1{,}202$ chunks.

\section{Dilution Analyses}
\label{app:dilution}

\begin{figure*}[t]
\centering
\includegraphics[width=0.82\textwidth]{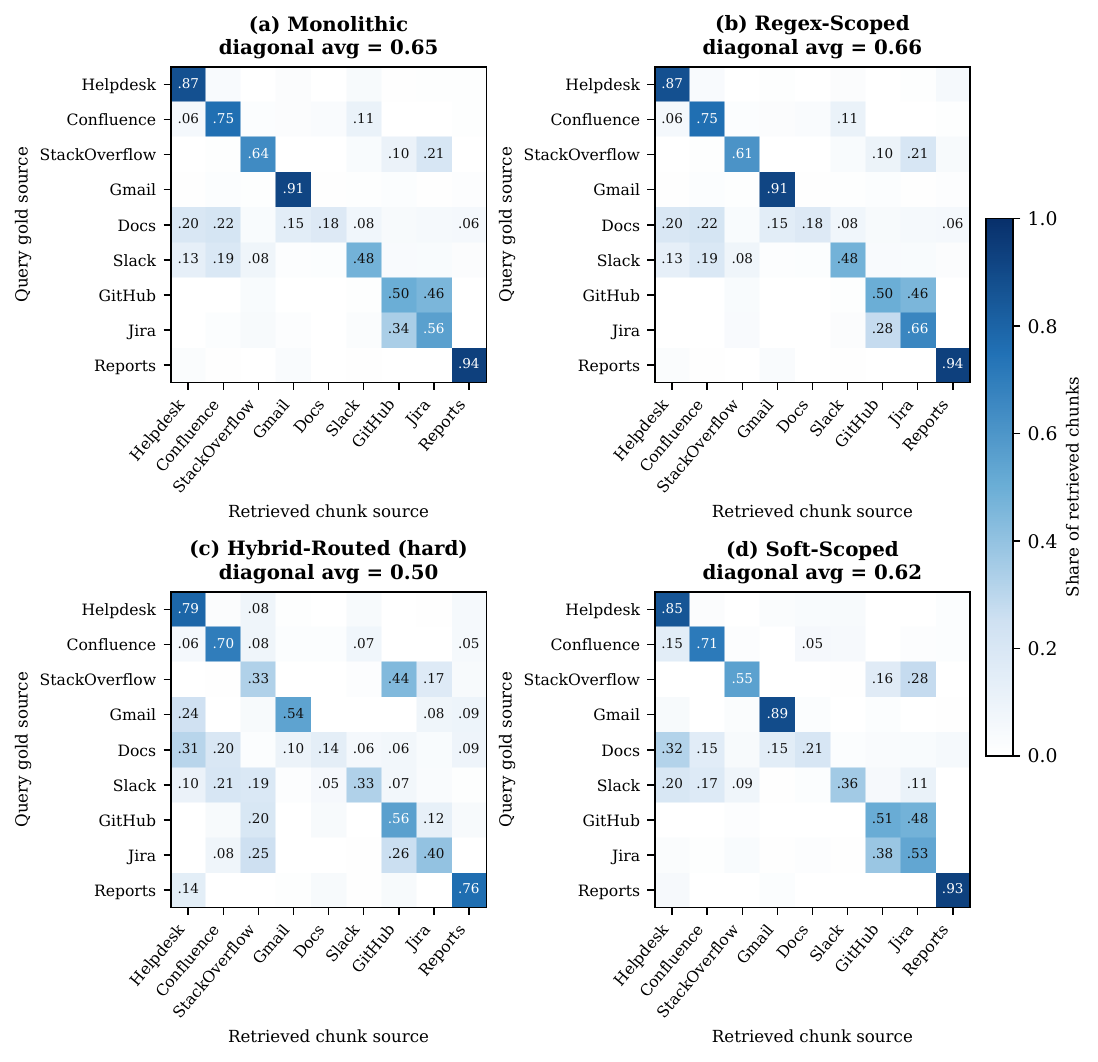}
\caption{Share of retrieved chunks by source, given the query's gold source (Composite-9, top-$15$). The diagonal is the share from the correct source; hard scoping lowers it (\hybridname{}, $0.50$), while Soft-Scoped stays close to Monolithic ($0.62$ vs.\ $0.65$).}
\label{fig:retrieval_confusion}
\end{figure*}

\subsection{Per-Query Regression}
\label{app:density_debias}

This analysis links category size to answer quality rather than retrieval. Each of $147$ queries judged under both monolithic and scoped retrieval contributes $\Delta_q = \mathrm{Corr}_{\text{scoped}}(q) - \mathrm{Corr}_{\text{global}}(q) \in \{-1,0,+1\}$, which we regress on the log chunk count $N_c$ of the query's category:
\[
\Delta_q = \beta_0 + \beta_1 \log_{10} N_c + \varepsilon_q .
\]
On the BGE-M3 + Qwen stack, $\hat\beta_1 = -0.217$ (SE $0.075$, $p{=}0.004$); on the Gemini production stack, $\hat\beta_1 = -0.159$ ($p{=}0.089$). Both slopes are negative: scoping improves correctness more for queries in smaller categories.

\subsection{Embedding-Space Views}
\label{app:tsne}

\begin{figure}[!htbp]
\centering
\includegraphics[width=\columnwidth]{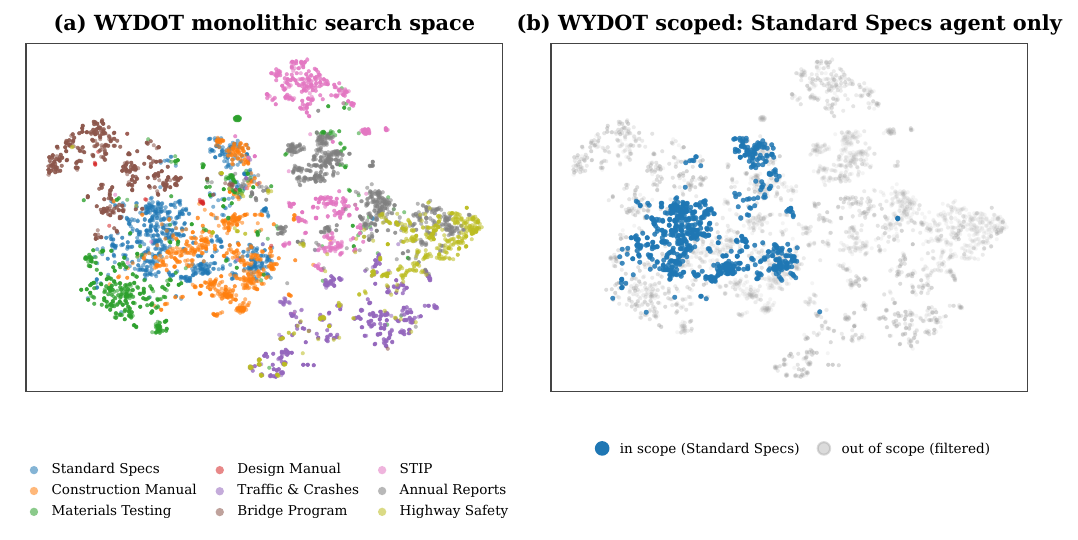}
\caption{WYDOT, t-SNE of up to $800$ chunks per category. (a) Monolithic: small categories scatter around large ones. (b) Standard Specs scope: one coherent region.}
\label{fig:embedding_viz}
\end{figure}

\begin{figure}[!htbp]
\centering
\includegraphics[width=\columnwidth]{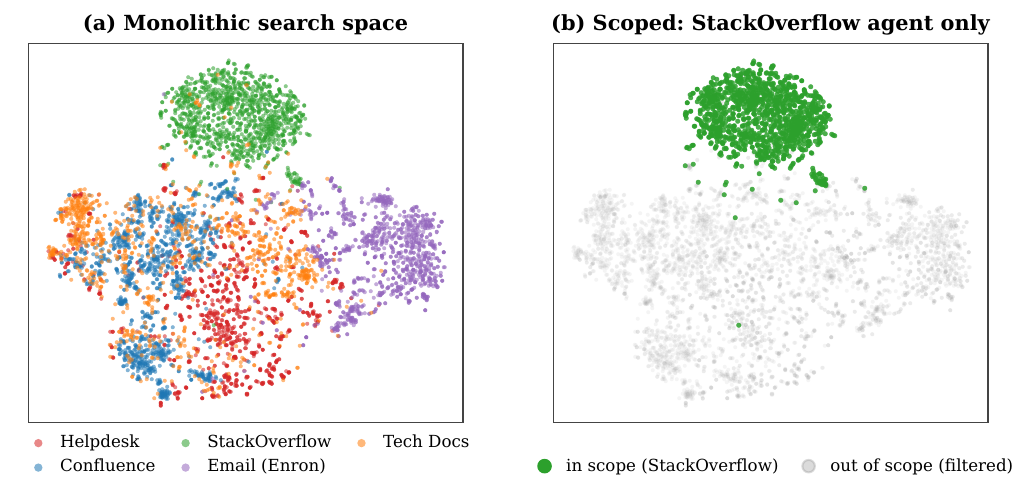}
\caption{Composite-9, t-SNE of up to $1{,}200$ chunks per source ($10{,}442$ total). (a) Monolithic: several sources overlap. (b) StackOverflow scope: one source remains.}
\label{fig:embedding_viz_composite}
\end{figure}

Figures~\ref{fig:embedding_viz} and~\ref{fig:embedding_viz_composite} show embedding-space dilution for WYDOT and Composite-9. Small categories sit at the edges of large, dense ones. A global nearest-neighbor search therefore favors the dense neighbors and misses chunks from the small category. A scope filter removes this bias by keeping only chunks with the target label.

Both views project BGE-M3 chunk embeddings, reduced to $50$ dimensions with PCA, using t-SNE with cosine distance, perplexity $35$, and seed $0$. Panel~(a) colors every sampled chunk by source.  Panel~(b) retains the coordinates of panel~(a) and grays out chunks outside the highlighted scope, which is what the metadata filter removes.

\paragraph{Retrieval-source confusion.} Figure~\ref{fig:retrieval_confusion} shows, for each gold source in Composite-9, which of the nine sources the top $15$ retrieved chunks come from ($225$ queries, canonical configuration). The diagonal average is the share of chunks from the correct source. Hard scoping lowers it from $0.65$ (Monolithic) to $0.50$ because a misrouted query searches the wrong source; Soft-Scoped keeps it close to Monolithic ($0.62$) because the global list always participates.

\section{Implementation}
\label{app:implementation}

\subsection{Production System}

\paragraph{Knowledge graph:} Documents were parsed with PyPDF2 and pdfplumber, split at DIVISION, SECTION, and CHAPTER headings, and chunked with LangChain's \texttt{SemanticChunker}. Chunks are embedded with Gemini Embedding ($3{,}072$-d) for the production stack and BGE-M3 ($1024$-d) for the open-source stack. The Neo4j AuraDB graph has $152{,}231$ nodes and $338{,}569$ relationships.

\paragraph{Agents and search:} Eight domain agents and one general agent share one interface: each domain agent applies its document\_series filter (App.~\ref{app:filters}) and exposes search, get\_section, and compare\_versions tools, while the general agent searches the full index. Within a scope, vector search (HNSW, cosine) and BM25 full-text search run in parallel, and duplicates are merged in favor of the vector hit. The production system runs as a web application on Google Cloud Run.

\subsection{Open-Source Backbones}
\label{app:oss_config}

\paragraph{Models:} Qwen2.5-7B-Instruct uses its native tool-calling template, whose \texttt{<tool\_call>} blocks we parse back to the orchestrator. Llama-3-8B-Instruct has no native template, so we use the same Hermes-style convention and list the tools in the system prompt.

\paragraph{Embedder:} BGE-M3 produces $1024$-d $L_2$-normalized vectors. The FAISS store uses exact search (\texttt{IndexFlatIP}), so dilution cannot come from approximate nearest-neighbor errors. As a sanity check, BGE-M3 reaches Recall@10 $0.961$ on $1{,}000$ BEIR MS-MARCO dev queries over a $200$k-passage subset, which is higher than on the full corpus \citep{bge_m3}. Decoding settings and hardware are in App.~\ref{app:reproducibility}.

\section{Metrics, Prompts, and Judging}
\label{app:prompts}

\subsection{Metrics}
\label{app:metrics}

\paragraph{Retrieval:} A retrieved chunk is relevant if it belongs to the query's labeled target scope. \textbf{P@10} is the share of the top $10$ chunks that are relevant. \textbf{R@10} is the share of a system's relevant retrieved chunks that rank in the top $10$, so it is lower for systems that retrieve many chunks, such as \systemname{}. \textbf{Gold-document variants} count a chunk as relevant only if it comes from the query's gold documents, which only $32$ WYDOT queries have.

\paragraph{Answer quality and cost:} Correctness and faithfulness are binary judgments averaged over queries (App.~\ref{app:judge}). Cost is reported as the median (p50) and 95th-percentile (p95) latency, mean tokens, and mean LLM calls per query.

\subsection{Routing Prompt}
\label{app:routing_prompt}

The LLM router assigns each query one of ten categories: Standard Specs, Construction Manual, Materials Testing, Design Manual, Traffic Crashes, STIP, Annual Report, Bridge Program, Highway Safety, and General (cross-domain or unclear). It also extracts a year and a document series when the query names them. Traffic Crashes and Highway Safety both map to the safety agent.

\subsection{Answer Prompts}
\label{app:answer_prompt}

The \emph{deployment} prompt (a) is what the production WYDOT system runs; it produced the Gemini production-stack results (Table~\ref{tab:retrieval}). All shared-prompt results (Table~\ref{tab:cross_domain}) use the corpus-neutral prompt (b) for every system, which isolates the retrieval architecture from prompt wording. Results labeled \emph{original answer prompt} predate this control: Monolithic used a bare prompt and the scoped systems used a persona prompt (App.~\ref{app:prompt_control}).

\medskip\noindent\textbf{(a) Deployment prompt (production WYDOT stack, abridged).}\par\nobreak\smallskip

\begin{lstlisting}[basicstyle=\ttfamily\scriptsize]
You are the WYDOT Knowledge Graph Assistant.
Answer questions about the Wyoming Department of
Transportation using a knowledge graph with
1,128 documents and 88,907 text chunks.

You must ALWAYS call at least one search tool
for EVERY query.
...
CITATION RULES:
- Reference sources as [Source 1], [Source 2].
- Include document title, section, and year.
...
- You can make up to 5 tool calls per query.
- Always ground answers in retrieved content.
\end{lstlisting}

\medskip\noindent\textbf{(b) Shared corpus-neutral prompt.}\par\nobreak\smallskip
\begin{lstlisting}[basicstyle=\ttfamily\scriptsize]
Answer the question using ONLY the retrieved
sources. Cite as [Source 1], [Source 2], etc.

Retrieved sources:
{context}

Question: {query}
Answer:
\end{lstlisting}

For the orchestrator and the ReAct baseline, the corpus persona is removed in the same way, and the tool-calling and loop instructions are kept verbatim. MA-RAG and LangChain ReAct keep their original prompts, and SCOUT-RAG uses our reimplementation's prompts; we equalize all three at the judge instead (App.~\ref{app:external_marag_scout}).

\subsection{Judge Prompts and Rubric}
\label{app:judge}

We use Qwen2.5-7B-Instruct as the judge. Each answer receives two greedy (temperature $0$) calls, each returning a binary score and a one-sentence reason. \textbf{Correctness} is $1$ if the answer conveys the same factual content as the reference, allowing for paraphrase. \textbf{Faithfulness} is $1$ if every factual claim in the answer is supported by at least one retrieved chunk shown to the judge (the top $15$, each truncated to $800$ characters); citation markers are not required. System-level Corr and Faith are the means of these binary scores.

\medskip\noindent\textbf{Correctness prompt.}\par\nobreak\smallskip
\begin{lstlisting}[basicstyle=\ttfamily\scriptsize]
You are evaluating a model-generated answer
against a reference answer.

Question: {query}
Reference answer: {reference}
Model answer: {answer}

Does the MODEL ANSWER convey the same factual
content as the REFERENCE ANSWER, allowing for
paraphrasing? Reply with a JSON object only:
{"correct": 1 or 0, "reason": "<one short sentence>"}
\end{lstlisting}

\medskip\noindent\textbf{Faithfulness prompt.}\par\nobreak\smallskip
\begin{lstlisting}[basicstyle=\ttfamily\scriptsize]
You are evaluating whether a model-generated
answer is grounded in the retrieved sources.

Retrieved sources:
{sources}

Model answer: {answer}

Is EVERY factual claim in the MODEL ANSWER
supported by at least one source above?
(Citations like [Source 1] don't need to be
present; just check whether the claims appear
in the sources.)

Reply with a JSON object only:
{"faithful": 1 or 0, "reason": "<one short sentence>"}
\end{lstlisting}

\subsection{Shared-Prompt Control}
\label{app:prompt_control}

Answer-prompt wording can confound scoped-versus-monolithic comparisons. In the original configuration, Monolithic used a bare prompt, while the scoped systems used a WYDOT persona that also asked the model to note gaps in the evidence. An audit of MultiHop exposed the effect. Monolithic and Soft-Scoped retrieved nearly the same chunks, yet their correctness differed by up to $19$ points, and their answer lengths followed the prompt ($667$ vs.\ $866$ characters). We therefore run every system on one corpus-neutral prompt, and the orchestrator keeps only its tool instructions. Table~\ref{tab:cross_domain} uses this control; Table~\ref{tab:wydot_extbaselines} does not. In the control run, the MultiHop gap shrinks from $\Delta{=}{-}0.094$ ($p{<}0.001$) to $\Delta{=}{-}0.010$ ($p{=}0.56$).

\section{External Baselines}
\label{app:external_baselines}

The external baselines below use the shared answer prompt. Three exceptions exist: MA-RAG and LangChain ReAct retain their original prompts, and SCOUT-RAG uses the prompts from our reimplementation. The baselines do not change the main conclusion. The multi-agent baselines are less faithful than Monolithic on every corpus and less accurate on most.

\subsection{Retrieval and Agent Baselines on Composite-9}
\label{app:composite_extbaselines}

\begin{itemize}[leftmargin=*,topsep=2pt,itemsep=1pt]
\item \textbf{BM25} \citep{robertson2009bm25}: the top $10$ chunks go to Qwen-7B.
\item \textbf{ColBERTv2} \citep{santhanam2022colbertv2}: the colbertv2.0 checkpoint with PLAID indexing. The top $10$ chunks go to Qwen-7B.
\item \textbf{ColBERTv2-scoped}: the same index, filtered to each query's \emph{gold} source. It is an oracle upper bound, not a deployable system.
\item \textbf{Custom ReAct} \citep{yao2023react}: our ReAct loop over the same nine tools (App.~\ref{app:react}).
\item \textbf{LangChain ReAct} \citep{chase2023langchain}: LangChain's ReAct agent over the same nine tools. It uses the same Qwen-7B model and decoding settings.
\end{itemize}

With oracle routing, scoping also helps a late-interaction retriever. ColBERTv2 restricted to the gold source reaches $57.3\%$ correctness and $.547$ faithfulness. Unscoped, it reaches $52.9\%$ and $.484$ ($n{=}225$; Table~\ref{tab:external_baselines}).

\begin{table}[!htbp]
\centering
\footnotesize
\setlength{\tabcolsep}{3pt}
\begin{tabular}{l r r r r}
\toprule
\textbf{Baseline} & \textbf{p50 (s)} & \textbf{p95 (s)} & \textbf{Corr\%} & \textbf{Faith} \\
\midrule
BM25 $+$ Qwen        & $2.17$ & $7.65$  & $74.2$ & $.65$ \\
ColBERTv2 $+$ Qwen   & $2.15$ & $8.79$  & $52.9$ & $.48$ \\
ColBERTv2-scoped$^{\ddagger}$ & $1.61$ & $8.64$  & $57.3$ & $.55$ \\
Custom ReAct ($n{=}222$) & $7.09$ & $19.42$ & $71.6$ & $.32$ \\
\midrule
\multicolumn{5}{l}{\emph{Original protocol ($n{=}50$; not re-run, see caption)}} \\
LangChain ReAct      & $12.42$ & $33.70$ & $48.0$ & $.10$ \\
\bottomrule
\end{tabular}
\caption{External baselines on Composite-9. Top block: shared answer prompt ($n{=}225$). Bottom block: LangChain ReAct under the original protocol ($n{=}50$), since its prompts are internal to the framework. ${}^{\ddagger}$Oracle: routes on the gold source.}
\label{tab:external_baselines}
\end{table}

\subsection{MA-RAG and SCOUT-RAG}
\label{app:external_marag_scout}

\paragraph{MA-RAG (port).} MA-RAG \citep{nguyen2025marag} chains four agents: Planner, Step-Definer, Extractor, and QA. Its public code supports only OpenAI models, so we port its prompts verbatim to our Qwen-2.5-7B and BGE-M3 stack. Outputs use JSON prompting with tolerant parsing. Retrieval is global, because MA-RAG does not use organizational metadata.

\paragraph{SCOUT-RAG (reimplementation).} SCOUT-RAG \citep{li2026scoutrag} runs four cooperative agents (DRAA, PAGA, OASA, and AQAA) in a refinement loop. The paper does not release its prompts, fusion functions, or stopping threshold, so we follow its Algorithm~1 and supply these parts ourselves ($\epsilon{=}0.05$). Each of our scopes becomes one SCOUT-RAG domain. Both baselines use the same LLM and retriever as our systems; only the coordination differs.

\paragraph{Equalized faithfulness (${}^{\ddagger}$).} We judge both baselines with the chunk-text database loaded, exactly like every other system; without it, every answer scores $0$. Their faithfulness is below Monolithic on every corpus: MA-RAG $0.070$, $0.378$, and $0.100$ on WYDOT, Composite-9, and MultiHop-RAG, and SCOUT-RAG $0.097$, $0.347$, and $0.220$. Their correctness is also lower on Composite-9 ($44.4\%$ and $66.7\%$, against $78.2\%$) and MultiHop-RAG ($29.8\%$ and $55.2\%$, against $69.0\%$). On WYDOT's BGE-M3 store, MA-RAG is below Monolithic and SCOUT-RAG slightly above ($11.0\%$ and $24.1\%$, against $21.6\%$; Table~\ref{tab:singlecall}). They also cost more: MA-RAG makes $22$--$26$ LLM calls per query and SCOUT-RAG $8$--$10.4$, against one for Monolithic.

\subsection{HotpotQA}
\label{app:hotpot_span}

We also replicate on HotpotQA-distractor ($4$ alphabetic topic buckets, $n{=}2{,}000$) and report its span-level metrics so the numbers can be compared with prior HotpotQA work (Table~\ref{tab:hotpot_emf1}). Strict EM is near zero because RAG answers are paragraph-length, so we add two long-form variants: Window-EM (a contiguous token-window match) and Contains (substring match on the normalized prediction).

\begin{table}[!htbp]
\small
\centering
\setlength{\tabcolsep}{4pt}
\begin{tabular}{lcccc}
\toprule
\textbf{System} & EM & F1 & winEM & Contains \\
\midrule
BM25 $+$ Qwen     & $.000$ & $.068$ & $.421$ & $.429$ \\
ColBERTv2 $+$ Qwen & $.000$ & $.072$ & $.446$ & $.454$ \\
Monolithic        & $.000$ & $.071$ & $.448$ & $.455$ \\
Regex-Scoped      & $.000$ & $.071$ & $\mathbf{.453}$ & $\mathbf{.458}$ \\
Soft-Scoped       & $.000$ & $.069$ & $.438$ & $.444$ \\
\hybridname{}     & $.000$ & $.042$ & $.255$ & $.272$ \\
\systemname{}     & $.000$ & $.038$ & $.316$ & $.338$ \\
Custom ReAct      & $.000$ & $\mathbf{.091}$ & $.382$ & $.401$ \\
\midrule
\multicolumn{5}{l}{\emph{Original protocol; not re-run (see caption)}} \\
LangChain ReAct   & $.000$ & $.059$ & $.264$ & $.277$ \\
\bottomrule
\end{tabular}
\caption{HotpotQA-distractor span-level metrics. Top block: shared answer prompt ($n{=}2{,}000$). Bottom block: LangChain ReAct under the original protocol, since its prompts are internal to the framework.}
\label{tab:hotpot_emf1}
\end{table}

\section{Routing and Scope Discovery}
\label{app:routing}

\subsection{Router Variants}
\label{app:router_variants}

We compare three routers on the $155$ labeled WYDOT queries ($5$-fold stratified cross-validation, $10$ query categories). \textbf{R0} is the production regex. \textbf{R1} is a class-balanced logistic regression over word $1$--$2$-gram TF--IDF features. \textbf{R2} is a logistic regression probe on the BGE-M3 query embedding. R2 improves accuracy by $28.4$ points and weighted $F_1$ by $24.4$ points over R0 (Table~\ref{tab:router_variants}). End-to-end, \textsc{R2-Routed} gives the best WYDOT correctness and Recall@10 in the original-prompt block (Table~\ref{tab:singlecall}), but under the shared prompt, it does not beat Monolithic (Table~\ref{tab:cross_domain}). A confidence-gated fallback to global retrieval lowers correctness steadily: $30.3\%$ at $\tau{=}0$, $25.3\%$ at $\tau{=}0.3$ ($68\%$ fallback), and $21.7\%$ as $\tau{\to}1$, close to Monolithic ($21.6\%$).

\begin{table}[!htbp]
\small
\centering
\setlength{\tabcolsep}{3pt}
\begin{tabular}{lccc}
\toprule
\textbf{Router} & \textbf{Acc}\,$_{95\%\,CI}$ & \textbf{Top-2} & \textbf{$F_1$}\,$_{95\%\,CI}$ \\
\midrule
R0 Regex   & $.471$                & $.471^\ddagger$ & $.497$ \\
R1 TF-IDF  & $.66_{.57,.74}$       & $.78{\pm}.11$  & $.63_{.52,.72}$ \\
R2 BGE-LR  & $\mathbf{.76}_{.72,.79}$ & $\mathbf{.86}{\pm}.03$ & $\mathbf{.74}_{.70,.79}$ \\
\bottomrule
\end{tabular}
\caption{Router accuracy ($n{=}155$, $5$-fold CV). R0 is single-label deterministic, so Top-$2$ equals Top-$1$ ($^\ddagger$).}
\label{tab:router_variants}
\end{table}

\paragraph{Retrieval budget.} Scoping's recall cost shrinks as the budget grows. On the $32$ WYDOT queries with gold documents (BGE-M3 store), raising $k$ from $3$ to $15$ lifts Regex-Scoped recall from $.156$ to $.312$, compared to $.156$ to $.219$ for Monolithic. \systemname{} plateaus ($.226$ to $.258$), so its recall loss is not recovered by a larger budget.

\subsection{Scope Discovery}
\label{app:discovery}

We average the BGE-M3 chunk embeddings of each document, normalize them, and cluster documents with $k$-means ($k{=}11$, seed $42$). Each chunk takes its document's cluster. Excluding the catch-all General series, chunk-weighted purity against the hand labels is $0.88$, and discovery splits General into five coherent sub-scopes. Document-level NMI is low ($0.27$), because $82\%$ of WYDOT documents are General. Discovery also works on CDOT, Caltrans, and Composite-9 (NMI $0.42$, $0.28$, and $0.65$).

Trained with identical folds and seed, the R2 router reaches $5$-fold accuracy of $0.787$ on hand labels and $0.748$ on discovered labels. A query's discovered label is the cluster of its gold document or, when it has none, the cluster that holds most chunks of its category. Both label sets are agent scopes, not the $10$ query categories of Table~\ref{tab:router_variants}, so these accuracies differ from its $.76$. Granularity cannot be chosen this way: FDOT, for example, has low document\_series alignment yet strong dilution along that axis.

\section{Paradox Ablations}
\label{app:paradox_ablations}

This appendix gives the full results behind Table~\ref{tab:paradox_causes}. The reranking, sparse-retrieval, and index comparisons use the original query subsets and answer prompt. Each comparison involves two arms on the same queries, so the prompt cancels within a row; however, absolute values are not comparable with Table~\ref{tab:cross_domain}.

\subsection{Within-Scope Reranking}
\label{app:rerank}

We rerank the top $30$ bi-encoder candidates with the bge-reranker-v2-m3 cross-encoder and keep the top $10$. Reranking raises faithfulness by $0.08$ (Monolithic) and $0.11$ (\hybridname{}) on Composite-9. It also helps Monolithic on the other public corpora, but its effect on \hybridname{} is mixed, and it hurts both systems on WYDOT (Table~\ref{tab:rerank}). These tests cover Monolithic and \hybridname{}, not \systemname{}. On WYDOT, the bi-encoder already ranks chunks by the section, year, and version the answer needs. The cross-encoder, which scores relevance in the style of monoT5~\citep{nogueira2020monoT5}, instead promotes on-topic chunks from the wrong year or version.

\begin{table}[!htbp]
\small
\centering
\setlength{\tabcolsep}{3pt}
\begin{tabular}{lccc}
\toprule
\textbf{Corpus / System} & $\Delta$Faith & $\Delta$Corr & $\Delta$R@10 \\
\midrule
\multicolumn{4}{l}{\textit{Monolithic}} \\
\quad WYDOT          & $-.065$ & $-.005$ & --- \\
\quad Composite-9    & $+.080$ & $+.040$ & $+.020$ \\
\quad MultiHop-RAG   & $+.064$ & $+.034$ & $+.005$ \\
\quad FinanceBench   & $+.021$ & $+.042$ & $+.035$ \\
\quad MMLU-Pro       & $+.026$ & $+.006$ & $+.002$ \\
\quad NQ-Open        & $+.026$ & $+.026$ & --- \\
\midrule
\multicolumn{4}{l}{\textit{\hybridname{}}} \\
\quad WYDOT          & $-.051$ & $-.011$ & --- \\
\quad Composite-9    & $+.114$ & $-.057$ & $+.029$ \\
\quad MultiHop-RAG   & $+.050$ & $+.012$ & $+.005$ \\
\quad FinanceBench   & $-.021$ & $+.056$ & $+.035$ \\
\quad MMLU-Pro       & $-.004$ & $-.018$ & $+.002$ \\
\quad NQ-Open        & $-.016$ & $-.028$ & --- \\
\bottomrule
\end{tabular}
\caption{Change from reranking (rerank minus base, same queries and prompt). R@10 is omitted for WYDOT and NQ-Open, which have no gold passages. Composite-9 \hybridname{} deltas use the $35$ queries with a baseline run.}
\label{tab:rerank}
\end{table}

\subsection{Sparse Retrieval}
\label{app:splade}

Replacing dense BGE-M3 with the learned-sparse SPLADE model~\citep{geng2024opensearchsparse} raises Monolithic correctness on Composite-9 ($90.0{\to}94.0$) and MultiHop ($69.2{\to}72.6$) and lowers it on FinanceBench ($54.7{\to}43.2$; Table~\ref{tab:splade}). Neither retriever family wins systematically.

\begin{table}[!htbp]
\small
\centering
\setlength{\tabcolsep}{3pt}
\begin{tabular}{l@{\hspace{3pt}}lccc}
\toprule
\textbf{Corpus} & \textbf{System} & Retriever & Faith & Corr\% \\
\midrule
\multirow{4}{*}{Composite-9} & Mono     & BGE-M3 & $.760$ & $90.0$ \\
                              & Mono     & SPLADE & $\mathbf{.880}$ & $\mathbf{94.0}$ \\
                              & Hybrid-R & BGE-M3 & $.771$ & $85.7$ \\
                              & Hybrid-R & SPLADE & $\mathbf{.880}$ & $\mathbf{98.0}$ \\
\midrule
\multirow{4}{*}{MultiHop} & Mono     & BGE-M3 & $.460$ & $69.2$ \\
                          & Mono     & SPLADE & $.464$ & $\mathbf{72.6}$ \\
                          & Hybrid-R & BGE-M3 & $\mathbf{.484}$ & $\mathbf{61.0}$ \\
                          & Hybrid-R & SPLADE & $.474$ & $60.2$ \\
\midrule
\multirow{4}{*}{FinanceB.} & Mono     & BGE-M3 & $\mathbf{.320}$ & $\mathbf{54.7}$ \\
                           & Mono     & SPLADE & $.308$ & $43.2$ \\
                           & Hybrid-R & BGE-M3 & $\mathbf{.340}$ & $\mathbf{52.7}$ \\
                           & Hybrid-R & SPLADE & $.288$ & $48.6$ \\
\bottomrule
\end{tabular}
\caption{SPLADE vs.\ dense BGE-M3 for single-call systems (Qwen-2.5-7B). Neither retriever wins consistently. Compare within a row; values are not comparable with Table~\ref{tab:cross_domain}.}
\label{tab:splade}
\end{table}

\subsection{Index Implementation}
\label{app:infra_parity}\label{app:local_neo4j}\label{app:unified_faiss}

We re-ran the five non-WYDOT corpora with exact FAISS search and with a local Neo4j~$5.26$ HNSW index. The local copy keeps the production database untouched. We held everything else fixed: the BGE-M3 query embedding, the scope filter, and the Qwen-7B synthesizer. Correctness and R@10 rankings do not change between the two indexes. Two faithfulness orderings flip by small margins (Monolithic and \hybridname{} on MultiHop, and \systemname{} on FinanceBench). Every difference is at most $.094$, with a median of about $.01$ (Table~\ref{tab:neo4j_vs_faiss}).

\begin{table}[!htbp]
\centering
\footnotesize
\setlength{\tabcolsep}{3pt}
\resizebox{\columnwidth}{!}{%
\begin{tabular}{l@{\hspace{4pt}}l@{\hspace{4pt}}ccccccc}
\toprule
& & \multicolumn{2}{c}{Faith} & \multicolumn{2}{c}{Corr} & \multicolumn{2}{c}{R@10} & \\
\cmidrule(lr){3-4}\cmidrule(lr){5-6}\cmidrule(lr){7-8}
\textbf{Corpus} & \textbf{System} & F & N & F & N & F & N & $|\Delta|_\text{max}$ \\
\midrule
Composite-9 & Monolithic    & $.760$ & $.780$ & $90.0$ & $82.0$ & $.920$ & $.880$ & $.080$ \\
            & \hybridname{} & $.771$ & $.800$ & $85.7$ & $82.0$ & $.943$ & $.880$ & $.063$ \\
            & \systemname{} & $.740$ & $.680$ & $74.0$ & $70.0$ & $.740$ & $.800$ & $.060$ \\
\midrule
MultiHop    & Monolithic    & $.460$ & $.470$ & $69.2$ & $71.0$ & $.975$ & $.977$ & $.018$ \\
            & \hybridname{} & $.484$ & $.466$ & $61.0$ & $60.8$ & $.975$ & $.977$ & $.018$ \\
            & \systemname{} & $.434$ & $.368$ & $61.6$ & $61.2$ & $.903$ & $.903$ & $.066$ \\
\midrule
FinanceBench & Monolithic   & $.320$ & $.313$ & $54.7$ & $58.5$ & $.847$ & $.837$ & $.038$ \\
             & \hybridname{} & $.340$ & $.333$ & $52.7$ & $51.0$ & $.847$ & $.837$ & $.017$ \\
             & \systemname{} & $.320$ & $.340$ & $64.0$ & $60.7$ & $.807$ & $.713$ & $.094$ \\
\midrule
MMLU-Pro    & Monolithic    & $.356$ & $.376$ & $48.8$ & $48.6$ & $.998$ & $1.000$ & $.020$ \\
            & \hybridname{} & $.408$ & $.400$ & $51.8$ & $50.6$ & $.998$ & $1.000$ & $.012$ \\
            & \systemname{} & $.344$ & $.355$ & $46.0$ & $46.3$ & $.682$ & $.687$ & $.011$ \\
\midrule
NQ-Open & Monolithic    & $.410$ & $.404$ & $28.2$ & $29.0$ & --- & --- & $.008$ \\
        & \hybridname{} & $.412$ & $.426$ & $32.2$ & $31.2$ & --- & --- & $.014$ \\
        & \systemname{} & $.390$ & $.374$ & $40.6$ & $36.6$ & --- & --- & $.040$ \\
\bottomrule
\multicolumn{9}{l}{\scriptsize NQ has no gold-passage labels (answer-only), so R@10 is omitted.} \\
\end{tabular}}
\caption{FAISS (F) vs.\ Neo4j (N) on the same queries. Compare F with N within a row. Composite-9 uses $n{=}50$ ($n{=}35$ for \hybridname{}), so its values are not comparable with Table~\ref{tab:cross_domain}.}
\label{tab:neo4j_vs_faiss}
\end{table}

\paragraph{Unified-FAISS WYDOT:} We also re-embedded every production WYDOT chunk with BGE-M3 into a FAISS store ($93{,}879$ chunks). With Qwen-2.5-7B ($n{=}197$), Monolithic, Regex-Scoped, and \hybridname{} reach $36.0\%$, $35.0\%$, and $34.0\%$ correctness. These values differ from the Neo4j store (Table~\ref{tab:singlecall}). However, on both stores, the three systems stay within $3$ points of one another.

\subsection{Single-Call Synthesis and the WYDOT Neo4j Block}
\label{app:singlecall_full}

\textsc{SingleCall} keeps \systemname{}'s retrieval but synthesizes once over all retrieved chunks, which removes fragmentation across rounds. On Composite-9 under the shared prompt ($n{=}225$), \textsc{SingleCall} reaches $70.7\%$ correctness and $.600$ faithfulness, compared to $62.2\%$ and $.591$ for \systemname{} (\S\ref{sec:paradox}). Both stay below Monolithic ($78.2\%$ and $.773$). Table~\ref{tab:singlecall} provides the earlier WYDOT block on the BGE-M3 Neo4j store under the original answer prompt, including \textsc{R2-Routed} and \textsc{SingleCall}.

\begin{table}[!htbp]
\scriptsize
\centering
\setlength{\tabcolsep}{6pt}\renewcommand{\arraystretch}{0.92}
\caption{WYDOT on the BGE-M3 Neo4j store under the original answer prompt ($n{=}199$). R@10 is gold-document recall on the $32$ queries with gold documents. Values are not comparable with Table~\ref{tab:cross_domain}.}
\label{tab:singlecall}
\begin{tabular}{lccc}
\toprule
\textbf{System} & R@10 & Faith & Corr\% \\
\midrule
Monolithic     & $.188$ & $.347$ & $21.6$ \\
Regex-Sc.      & $.219$ & $.296$ & $24.6$ \\
\hybridname{}  & $.194$ & $.340$ & $21.8$ \\
R2-Routed      & $\mathbf{.375}$ & $.369$ & $\mathbf{30.3}$ \\
\systemname{}  & $.258$ & $\mathbf{.391}$ & $27.4$ \\
SingleCall     & $.188$ & $.221$ & $15.1$ \\
\bottomrule
\end{tabular}
\end{table}

\subsection{Backbones}
\label{app:backbones_full}

\paragraph{Cross-backbone WYDOT:} Table~\ref{tab:backbones_full} reports three metrics per system for each generator under the original answer prompt. These rows show how large the effect is, not which model classes are exempt: the largest collapses are commercial (Claude, GPT-5-mini), but so is the one backbone without a collapse (DeepSeek-V3, $n{=}44$), and the Qwen block reverses under the shared prompt (\S\ref{sec:paradox}).

\begin{table}[!htbp]
\small
\centering
\setlength{\tabcolsep}{2.5pt}
\begin{tabular}{l@{\hspace{3pt}}l@{\hspace{3pt}}rccc}
\toprule
\textbf{Backbone} & \textbf{System} & $n$ & Faith & Corr\% & R@10 \\
\midrule
\multirow{4}{*}{Qwen-7B}
 & Mono         & $193$ & $.352$ & $21.2$ & $.188$ \\
 & Regex-Sc.    & $193$ & $.306$ & $24.4$ & $.219$ \\
 & Hybrid-R.    & $193$ & $.342$ & $21.8$ & $.194$ \\
 & MASDR        & $193$ & $\mathbf{.394}$ & $\mathbf{27.5}$ & $\mathbf{.258}$ \\
\midrule
\multirow{4}{*}{Claude}
 & Mono         & $100$ & $.250$ & $\mathbf{24.0}$ & $.188$ \\
 & Regex-Sc.    & $100$ & $.250$ & $21.0$ & $.219$ \\
 & Hybrid-R.    & $100$ & $\mathbf{.270}$ & $21.0$ & $.250$ \\
 & MASDR        & $100$ & $.010$ & $8.0$ & $\mathbf{.500}$ \\
\midrule
\multirow{4}{*}{GPT-5m}
 & Mono         & $29$ & $.378$ & $17.2$ & $.188$ \\
 & Regex-Sc.    & $29$ & $\mathbf{.414}$ & $24.1$ & $.219$ \\
 & Hybrid-R.    & $29$ & $.310$ & $27.6$ & $.219$ \\
 & MASDR        & $29$ & $.241$ & $\mathbf{41.4}$ & $\mathbf{.280}$ \\
\midrule
\multirow{4}{*}{DeepSeek}
 & Mono         & $44$ & $.227$ & $45.5$ & $.188$ \\
 & Regex-Sc.    & $44$ & $.364$ & $56.8$ & $.219$ \\
 & Hybrid-R.    & $44$ & $\mathbf{.364}$ & $\mathbf{61.4}$ & $.281$ \\
 & MASDR        & $44$ & $.318$ & $52.3$ & $\mathbf{.469}$ \\
\bottomrule
\end{tabular}
\caption{WYDOT across backbones (original answer prompt, BGE-M3 Neo4j store). R@10 is gold-document recall on the $32$ queries with gold documents. Within each backbone, $n$ counts the queries answered by all four systems.}
\label{tab:backbones_full}
\end{table}

\paragraph{Llama-3-8B on public corpora:} We repeat four public corpora (MultiHop-RAG, FinanceBench, MMLU-Pro, NQ-Open) with Llama-3-8B-Instruct, which has no native tool-calling template. The retriever, judge, and sample sizes match the Qwen runs (Table~\ref{tab:llama_cross}). With this backbone, \systemname{}'s faithfulness collapses. On FinanceBench it drops from $.207$ (\hybridname{}) to $.000$, because \systemname{} fails to produce a parsable answer for any query; on MMLU-Pro it drops from $.162$ to $.096$. The single-call systems, in contrast, stay within $.06$ faithfulness and $7$ correctness points of one another on every corpus. \systemname{} therefore depends on a synthesizer with strong built-in tool calling, which Llama-3-8B lacks.

\begin{table}[!htbp]
\small
\centering
\setlength{\tabcolsep}{3pt}
\caption{Llama-3-8B cross-corpus replication. Same retriever, scope filter, and judge as the Qwen runs; only the synthesizer differs.}
\label{tab:llama_cross}
\begin{tabular}{l@{\hspace{4pt}}l@{\hspace{4pt}}rcc}
\toprule
\textbf{Corpus} & \textbf{System} & $n$ & Faith & Corr\% \\
\midrule
\multirow{4}{*}{MultiHop} & Mono     & $500$ & $.410$ & $\mathbf{55.6}$ \\
                          & Regex-Sc.& $500$ & $\mathbf{.420}$ & $50.0$ \\
                          & Hybrid-R.& $500$ & $.392$ & $48.6$ \\
                          & MASDR    & $500$ & $.130$ & $42.0$ \\
\midrule
\multirow{4}{*}{FinanceB.} & Mono    & $150$ & $.200$ & $\mathbf{23.3}$ \\
                           & Regex-Sc.& $150$ & $.173$ & $20.7$ \\
                           & Hybrid-R.& $150$ & $\mathbf{.207}$ & $\mathbf{23.3}$ \\
                           & MASDR    & $150$ & $.000$ & $0.0$ \\
\midrule
\multirow{4}{*}{MMLU-Pro} & Mono     & $500$ & $\mathbf{.210}$ & $\mathbf{26.8}$ \\
                          & Regex-Sc.& $500$ & $.150$ & $26.0$ \\
                          & Hybrid-R.& $500$ & $.162$ & $25.6$ \\
                          & MASDR    & $500$ & $.096$ & $18.8$ \\
\midrule
\multirow{4}{*}{NQ-Open}  & Mono     & $500$ & $\mathbf{.350}$ & $24.0$ \\
                          & Regex-Sc.& $500$ & $.334$ & $26.0$ \\
                          & Hybrid-R.& $500$ & $.336$ & $25.0$ \\
                          & MASDR    & $500$ & $.094$ & $\mathbf{36.0}$ \\
\bottomrule
\end{tabular}
\end{table}

\section{Cost and the ReAct Baseline}
\label{app:react}

Our ReAct baseline is an Action--Observation loop of at most $6$ rounds. In each round, the LLM reads the current scratchpad and either calls a tool or gives a final answer. If it reaches the round limit, it must write a final answer based on what it has gathered. It uses the same nine retrieval tools as \systemname{}, so any difference reflects control flow, not retrieval.

ReAct averages $5.50$ iterations per query on Llama-3-8B and $2.26$ on Qwen-7B (Table~\ref{tab:oss_efficiency}, Figure~\ref{fig:efficiency}). We attribute the gap to Qwen's native function-calling template, which lets the model commit to a tool call instead of reasoning around it.

\begin{table}[!htbp]
\centering
\footnotesize
\resizebox{\columnwidth}{!}{%
\begin{tabular}{l l r r r r}
\toprule
\textbf{LLM} & \textbf{System} & \textbf{p50 (s)} & \textbf{p95 (s)} & \textbf{tokens} & \textbf{calls} \\
\midrule
Qwen-7B & monolithic    & 6.2  & 19.2  & 11.3k & 1.00 \\
Qwen-7B & regex\_scoped & 7.6  & 19.9  & 10.7k & 1.00 \\
Qwen-7B & \hybridname{} & 6.3  & 20.2  & 10.8k & 1.44 \\
Qwen-7B & \systemname{} & 10.8 & 31.7  & 13.0k & 2.09 \\
Qwen-7B & ReAct         & 7.9  & 19.7  & 11.6k & 2.26 \\
\midrule
Llama-8B & monolithic    & 9.9  & 51.0  & 7.4k  & 1.00 \\
Llama-8B & regex\_scoped & 12.7 & 51.7  & 7.7k  & 1.00 \\
Llama-8B & \hybridname{} & 9.2  & 51.0  & 6.6k  & 1.46 \\
Llama-8B & \systemname{} & 23.0 & 62.1  & 11.4k & 2.08 \\
Llama-8B & ReAct         & 20.3 & 143.8 & 38.7k & 5.50 \\
\bottomrule
\end{tabular}}
\caption{Open-source replication on WYDOT $200$-query: per-system latency, token, and LLM-call budgets. ReAct on Llama-8B uses far more calls and tokens than on Qwen-7B. Measured under the original answer prompt.}
\label{tab:oss_efficiency}
\end{table}

\section{Case Studies}
\label{app:case_studies}\label{app:failure_case}

\paragraph{Case 1: a query that breaks Monolithic and \systemname{}.} \emph{``What are the safety considerations for crusher site inspections?''} belongs to the Construction Manual ($6.6$k chunks) but also contains safety vocabulary. Monolithic retrieval is pulled to the much larger Traffic~\&~Crashes category ($30.9$k chunks) and answers questions about Highway Safety Plans. \systemname{}'s router mislabels the query as Highway Safety, with the same result. \hybridname{}'s regex matches ``inspection'', scopes to the Construction Manual, and retrieves the correct \S7-3 crusher-site procedure.

\paragraph{Case 2: version comparison, where orchestration helps.} For \emph{``What changed in aggregate gradation between $2010$ and $2021$?''}, \systemname{} calls its compare\_versions tool, retrieves Section~703 from both editions, and produces a structured comparison. Single-call scoped search has no such tool, so this is where orchestration adds value.

\section{Statistics and Reproducibility}
\label{app:stats}\label{app:reproducibility}

\paragraph{Statistical tests and data.} Pairwise comparisons use paired permutation tests ($10{,}000$ permutations, $\alpha{=}0.05$). Router accuracies (Table~\ref{tab:router_variants}) carry bootstrap $95\%$ confidence intervals ($1{,}000$ resamples). We release the WYDOT suite with the code, and all public corpora use their published splits.

\paragraph{Hyperparameters.} Routing uses temperature $0.0$. Answer generation uses temperature $0.2$, top-$p{=}0.9$, and a $1024$-token cap. The scoped and monolithic systems retrieve the top $15$ chunks; BM25 and ColBERTv2 retrieve the top $10$; and reranking re-scores the top $30$ down to $10$. Neo4j HNSW uses its default graph-construction settings. The R2 router uses class-balanced logistic regression with up to $4{,}000$ iterations.

\paragraph{Hardware and runtime.} Inference uses HuggingFace Transformers. One WYDOT evaluation ($200$ queries $\times$ $5$ systems) takes about $40$ minutes on an L40S with Qwen-7B and $100$ minutes on an A30 with Llama-8B. The roughly $70{,}000$ judge calls took about $24$ L40S GPU-hours.

\paragraph{Seeds.} Seed $0$ fixes t-SNE, bootstrap intervals, and the R2 cross-validation splits. Seed $42$ fixes the query generator and $k$-means scope discovery. Routing and judging are greedy, but answer generation is sampled without a fixed seed. Regenerated answers will therefore differ slightly, but every table can be recomputed from the released judged logs.

\section{What We Do Not Claim}
\label{app:scope}

\begin{itemize}[leftmargin=*,topsep=2pt,itemsep=2pt]
\item \textbf{That scoping improves WYDOT correctness.} Under the shared prompt, no scoped system significantly beats Monolithic on WYDOT. Scoping's demonstrated value there is retrieval precision.
\item \textbf{That soft-scoping preserves faithfulness.} It never significantly lowers correctness, but it does lower faithfulness on Composite-9 and MMLU-Pro.
\item \textbf{That \systemname{} is generally better.} It beats Monolithic on every metric only in the original-prompt BGE-M3 block (Table~\ref{tab:singlecall}). Under the shared prompt, it has the lowest faithfulness of any system on WYDOT ($0.33$ vs.\ $0.49$).
\item \textbf{The size of the paradox.} The automatic judge overstates the gap compared with the human rater ($0.56$ vs.\ $0.17$), so we claim only its direction.
\end{itemize}

\end{document}